\documentclass[runningheads]{llncs}
\usepackage{graphicx}
\usepackage[dvipsnames]{xcolor, colortbl}
\usepackage{algorithm2e}
\usepackage{subfig}
\usepackage{array}
\usepackage{tikz}
\usepackage{amssymb,amscd,amsmath}
\usepackage{cleveref}
\usepackage[export]{adjustbox}
\usepackage{multicol}
\usepackage{multirow}
\usepackage{subfloat}

\usepackage{physics}
\definecolor{gray}{RGB}{208,208,208}
\usepackage{standalone}
\usepackage{import}
\usepackage{floatrow}
\DeclareMathOperator*{\argmax}{\arg\!\max}

\begin{document}
\title{survAIval: Survival Analysis \\with the Eyes of AI} %\thanks{Supported by organization x.}}

\author{Kamil Kowol\inst{1} \and
Stefan Bracke\inst{2} \and
Hanno Gottschalk\inst{3}}

\authorrunning{K. Kowol et al.}
\institute{School of Mathematics and Natural Sciences, IZMD, University of Wuppertal, Gaußstraße 20, Wuppertal, Germany \and
Chair of Reliability Engineering and Risk Analytics, IZMD, University of Wuppertal, Gaußstraße 20, Wuppertal, Germany \and
Institute of Mathematics, Technical University Berlin, Straße des 17. Juni 135, Berlin \\
\email{kowol@math.uni-wuppertal.de},
\email{bracke@uni-wuppertal.de},
\email{gottschalk@math.tu-berlin.de}
}

\maketitle              % typeset the header of the contribution
\begin{abstract}
In this study, we propose a novel approach to enrich the training data for automated driving by using a self-designed driving simulator and two human drivers to generate safety-critical corner cases in a short period of time, as already presented in~\cite{kowol22simulator}. Our results show that incorporating these corner cases during training improves the recognition of corner cases during testing, even though, they were recorded due to visual impairment. Using the corner case triggering pipeline developed in the previous work, we investigate the effectiveness of using expert models to overcome the domain gap due to different weather conditions and times of day, compared to a universal model from a development perspective. Our study reveals that expert models can provide significant benefits in terms of performance and efficiency, and can reduce the time and effort required for model training. Our results contribute to the progress of automated driving, providing a pathway for safer and more reliable autonomous vehicles on the road in the future.

\end{abstract}
\section{\uppercase{Introduction}}
\label{sec:introduction}
If automotive manufacturers want to put autonomous vehicles higher than level 2 on the road, they should ensure that safety-critical driving situations are registered and that a safe solution for all road users is found as quickly as possible. One way to achieve this is to provide a large amount of diverse data to the model during training to increase the robustness and performance of AI algorithms. However, large amounts of annotated data alone may not ensure safe operation in those rare situations where road users are exposed to significant risk. For this reason, we introduced the A-Eye method~\cite{kowol22simulator} to apply an accelerated testing strategy that exploits human risk perception to capture corner cases and thereby achieve performance improvements in safety-critical driving situations. To this end, a self-designed driving simulator was developed that detects safety-critical driving situations in real-time based on poor AI predictions. With the help of this driving simulator and a further driving campaign, the domain shift will be investigated on different weather domains. Closing the gap of domain shifts due to different weather conditions requires targeted data generation from multiple domains to achieve a good performance. Even if using more data and the best models leads to overcoming the domain gap, the question is whether this is the most efficient way from the manufacturer's point of view. In this regard, we investigate whether overcoming the domain gap in different weather conditions with specialized models works as well as or even better than a universal model in the sense that all weather modalities are covered during training.
This involves training a baseline model on sunny and daytime images, and then measuring in 600-second drives how long it takes for a corner case to occur in one of the following conditions: rain, fog or night. An expert model is then trained for each weather condition, which retrains the baseline model for that domain. Finally, a universal model is trained, which is exposed to all weather parameters during training. The expert and universal models are also tested using the same scheme as the baseline model to measure the duration of a corner case in case one occurs.

\paragraph*{Outline}
\Cref{sec:simulator} introduces the self-designed driving simulator with the software and hardware used, followed by a corner case definition. A corner case triggering pipeline is then presented and used in test field. 
\Cref{sec:survival} discusses the basics of survival analysis to evaluate the drives from the weather-driving campaign. Finally, we present our conclusions and give an outlook on future directions of research in \Cref{sec:conclusion}.

\section{\uppercase{Driving Simulator}}\label{sec:simulator}
There is an increased interest in human-in-the-loop (HITL) and machine learning approaches, where humans interact with machines to combine human and machine intelligence to solve a given problem~\cite{Wu2021hitlsurvey}. For this purpose simulators were used to improve AI systems by means of human experience or to study human behavior in field trials. We have therefore developed a test rig in which two human drivers can control a vehicle in real-time, with the visual output of a semantic segmentation network displayed on one driver's screen, while the other driver sees the untouched original image.

\begin{figure}[h!]
    \centering
    \captionsetup[subfigure]{labelformat=empty}
	\subfloat[View of the semantic driver (top) and the safety driver (bottom).]{\includegraphics[width=0.49\textwidth]  {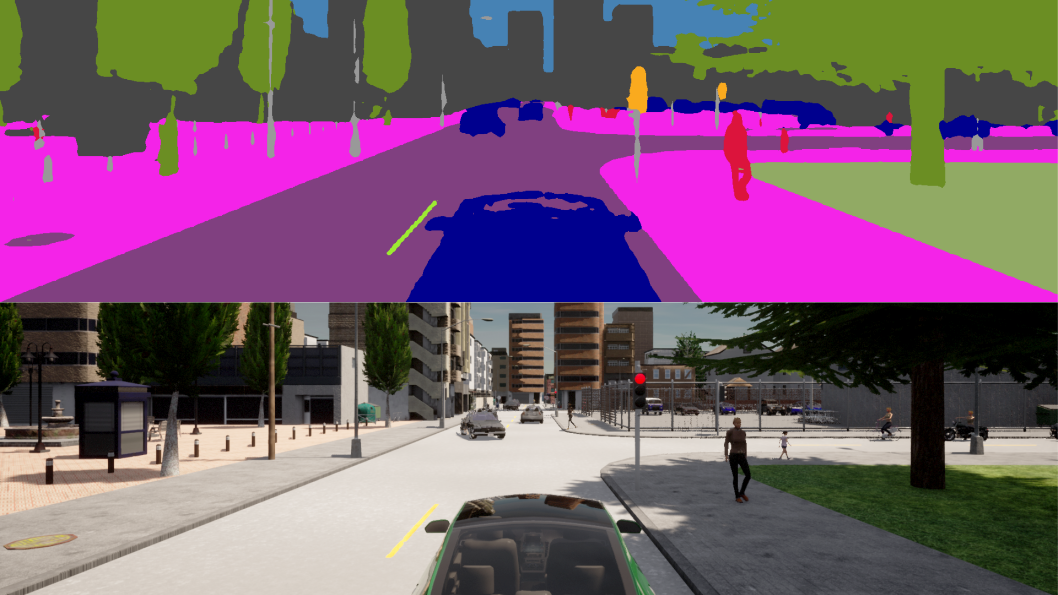}}~~
    \subfloat[Test rig including steering wheels, pedals, seats and screens.]{\includegraphics[width=0.49\textwidth, trim={0 8cm 0 18.5cm}, clip]	{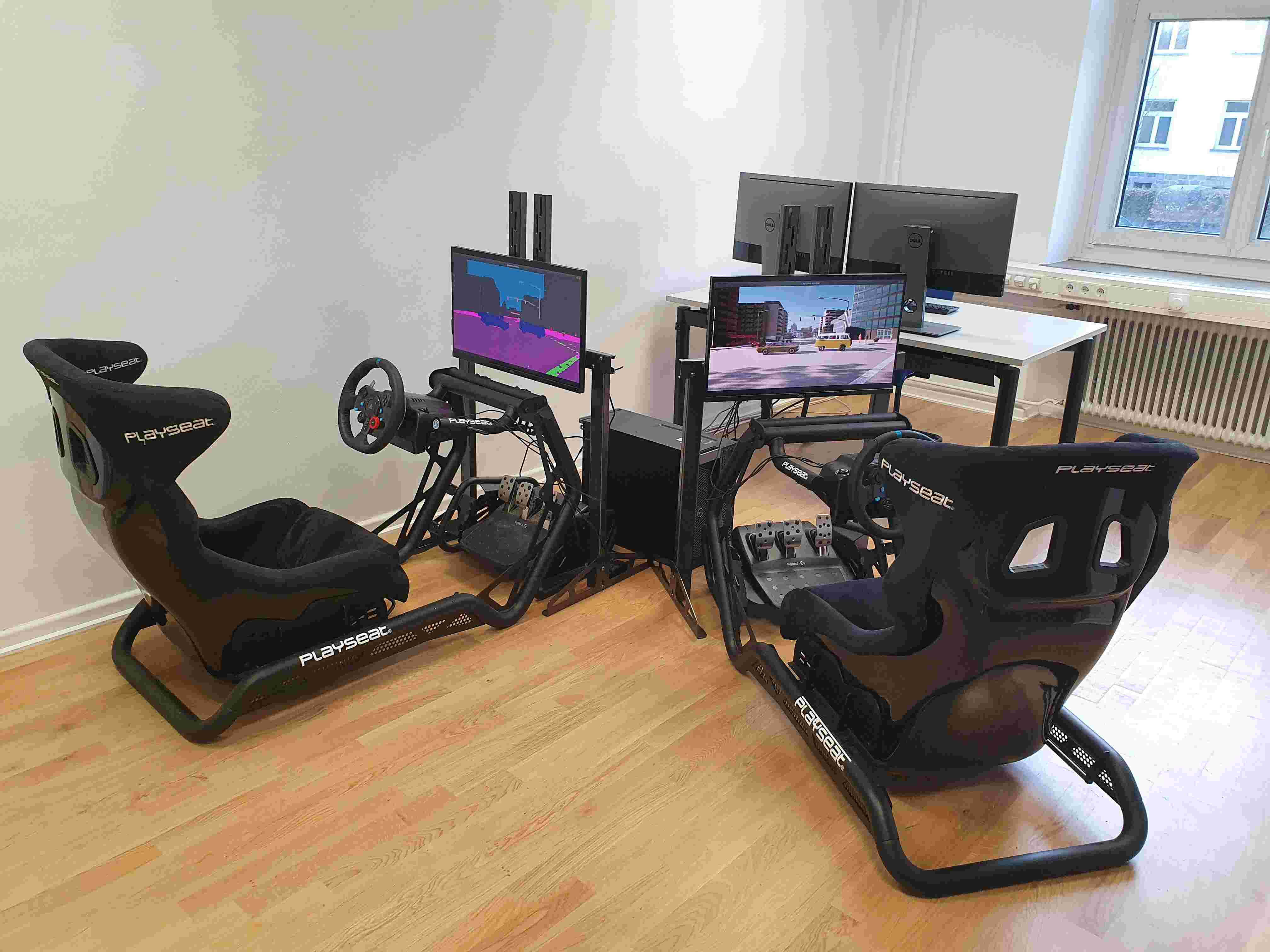}}~~
    \vspace{5pt}
    \caption{Harware and visual outputs of the A-Eye approach.}
    \label{fig:test_rig}
\end{figure}

By evaluating the same driving situation differently due to visual perception, we are able to find and save safety-critical driving situations in the shortest possible time, which can subsequently be used for training. This kind of targeted enrichment of training data with safety-critical driving situations is essential to increase the performance of AI algorithms. Since the generation of corner cases in the real world is not an option for safety reasons, generation remains in the synthetic world, where specific critical driving situations can be simulated and recorded. For this purpose, the autonomous driving simulator CARLA~\cite{dosovitskiy2017carla} is used. It is open-source software for data generation and/or testing of AI algorithms. It s various sensors to describe the scenes such as cameras, LiDAR as well as RADAR and provides ground truth data. CARLA is based on the Unreal Engine game engine~\cite{unrealengine}, which calculates and displays the behavior of the various road users with consideration of physics and thus enables realistic driving. In addition, with the Python API, the world can be modified and adapted to one's own use case. Therefore, we added another sensor, the inference sensor, to the script for manual control from the CARLA repository which evaluates the CARLA RGB images in real-time and outputs the prediction of a semantic segmentation network on the screen, see~\Cref{fig:test_rig}.
By connecting a control unit that s a steering wheel, pedals and a screen, it is possible to control a vehicle with 'the eyes of the AI' in the synthetic world of CARLA. Furthermore, we connected a second control unit with the same components to the simulator, so that it is possible to control the same vehicle with 2 different control units, see \Cref{fig:test_rig}. The second control unit, therefore, has control over the CARLA clear image and can intervene at any time. It always has priority and saves the past $3$ seconds of driving, which are buffered, on the hard disk. In order for the semantic driver to follow the traffic rules in CARLA, the script had to be modified to display the current traffic light phase in the top right corner and the speed in the top center.

\subsection{Test Rig}
The test rig consists of a workstation with dual Intel Xeon Gold 6258R as CPUs, 3x GPUs Quadro RTX 8000 and 1TB of RAM, which provides both high access speeds and sufficient memory swap calculations to meet the requirements of CARLA version $0.9.10$. The test rig also s $2$ driving seats, $2$ control units (steering wheel with pedals), one monitor for each control unit as well as two monitors for the control center. The control unit represents the interface between humans and machines. It enables the human to control a vehicle in CARLA freely via the steering wheel and the brake or throttle pedals. 
The device of choice was the Logitech G29~\cite{LogitechG29}, which is also pre-implemented in CARLA's control script and can therefore be used as a controller almost without any problems.

\subsection{Corner Cases}
When thinking about autonomous vehicles that move safely through traffic, it is necessary to perceive the environment correctly in order to provide safe driving. Especially the detection of atypical and dangerous situations is crucial for the safety of all road users. In order to improve the ability of today's models to handle such critical situations, datasets are required that allow for targeted training and, more importantly, testing with such critical situations. 
While there is no standard definition for the term \emph{corner case} in the context of autonomous driving, most definitions in the literature refer to rare but safety-critical driving situations. These scenarios can , for example, extreme weather conditions, as well as unexpected road obstacles that are uncommon but still need to be considered to ensure safe vehicle operation.

According to~\cite{Bolte2019}, a corner case for camera-based systems in the field of autonomous driving describes a \emph{"non-predictable relevant object/class in relevant location"}. This means that the unpredictable happens to moving objects (relevant class) interacting with each other on the road (crossing trajectories). Based on this definition, a corner case detection framework was presented to calculate a corner case score based on video sequences. The authors of~\cite{breitenstein2020cc} subsequently developed a systematization of corner cases, in which they divide corner cases into different levels and according to the degree of complexity. In addition, examples were given for each corner case level. This was also the basis for a subsequent publication with additional examples~\cite{Breitenstein2021}.
Due to the camera-based approach in the referenced works, a categorization of corner cases based on sensors was employed in~\cite{Heidecker2021}, which also included radar and LiDAR sensors. The authors defined four overarching layers - \emph{Sensor, Content, Temporal, and Method} - that incorporated the previously defined levels. As this definition is scientifically grounded and takes into account different sensor modalities, we would like to adopt it. 

While \emph{Sensor}, \emph{Content} and \emph{Temporal Layer} describe corner cases from the perspective of the human driver, the \emph{Method Layer} specifies corner cases in machine learning models due to lack of knowledge. Accordingly, epistemic uncertainty comes into play, which can be addressed by targeted data generation. Therefore, our focus is on this type of layer to increase safety. 

\subsection{Triggering Corner Cases}
Two test operators drive across the virtual world of CARLA and record scenes in our specially designed test rig, where one subject (safety driver) gets to see the original virtual image and the other (semantic driver) receives the output of the semantic segmentation network (see \Cref{fig:test_rig}). The test rig is equipped with controls such as steering wheels, pedals and car seats and connected to CARLA to create a simulated environment for realistic traffic participation.

\begin{figure}[h!]
	\centering
	\includegraphics[width=\linewidth, frame, trim={0 0 0 0}, clip]{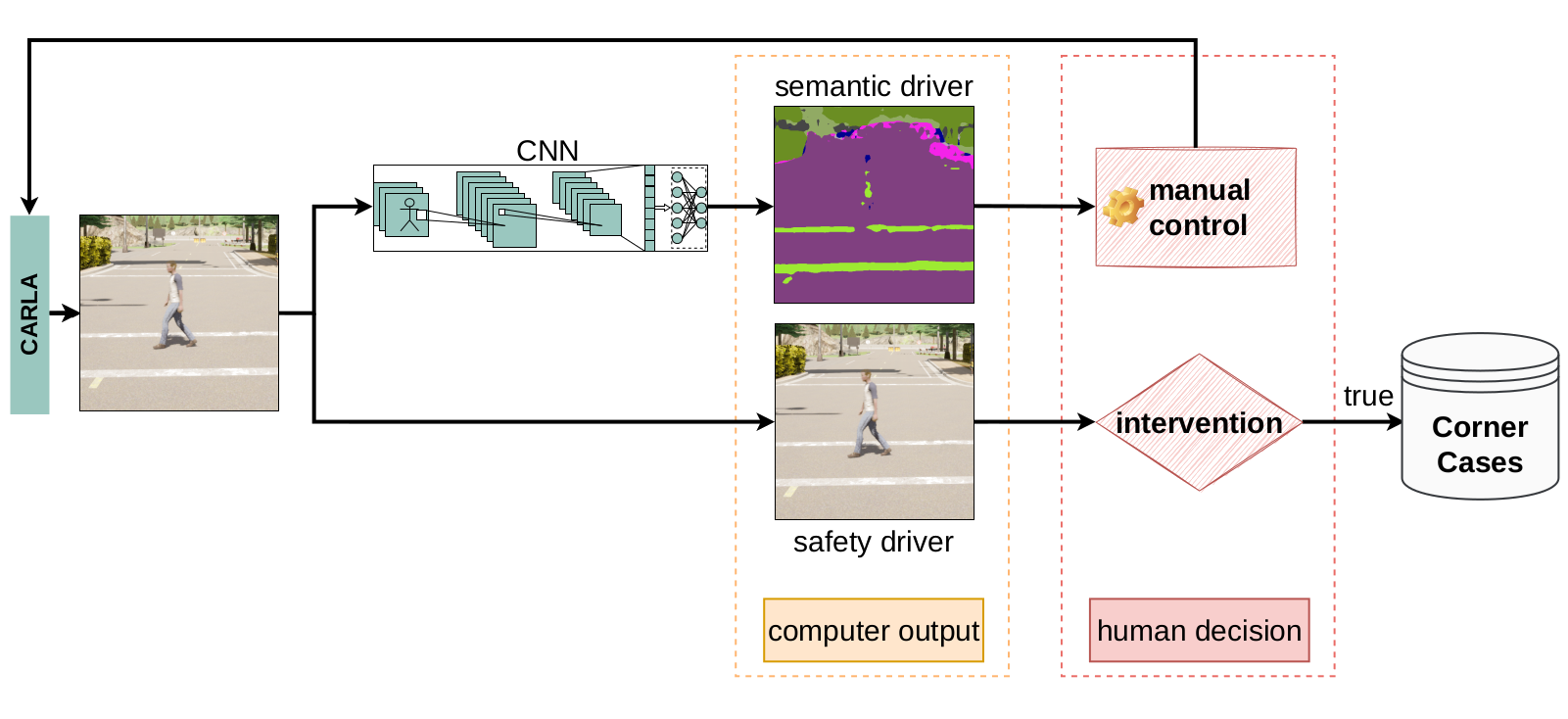}
	\caption[Flowchart for corner case triggering]{Two human subjects are able to control the ego-vehicle. Thereby, the semantic driver primarily controls the vehicle while following the traffic rules in the virtual world seeing only the output of the semantic segmentation network. The safety driver, who only sees the original image, takes on the role of a driving instructor and intervenes in the situation as soon as a dangerous situation arises. Intervening in the current situation indicates poor situation awareness of the segmentation network and represents a corner case, which simultaneously terminates the ride. The figure was already published in~\cite{kowol22simulator}.}
	\label{fig:flowchart}
\end{figure}

\begin{figure}[h!]
    \centering
	\captionsetup[subfloat]{labelformat=empty}
    \subfloat[][]{\includegraphics[width=0.49\textwidth, trim={5cm 210px 5cm 60px}, clip]{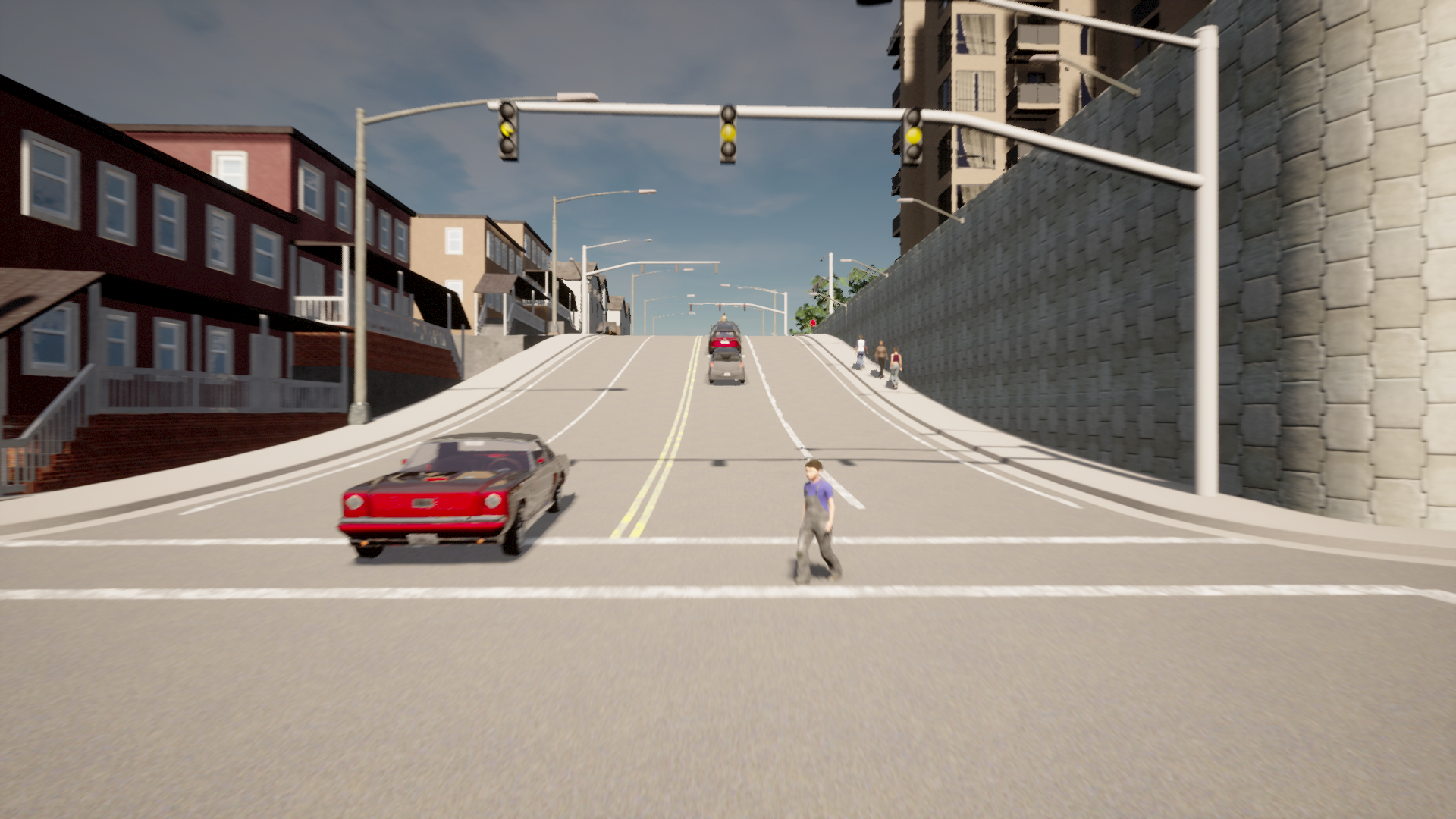}}~~
    \subfloat[][]{\includegraphics[width=0.49\textwidth, trim={5cm 150px 5cm 0}, clip]{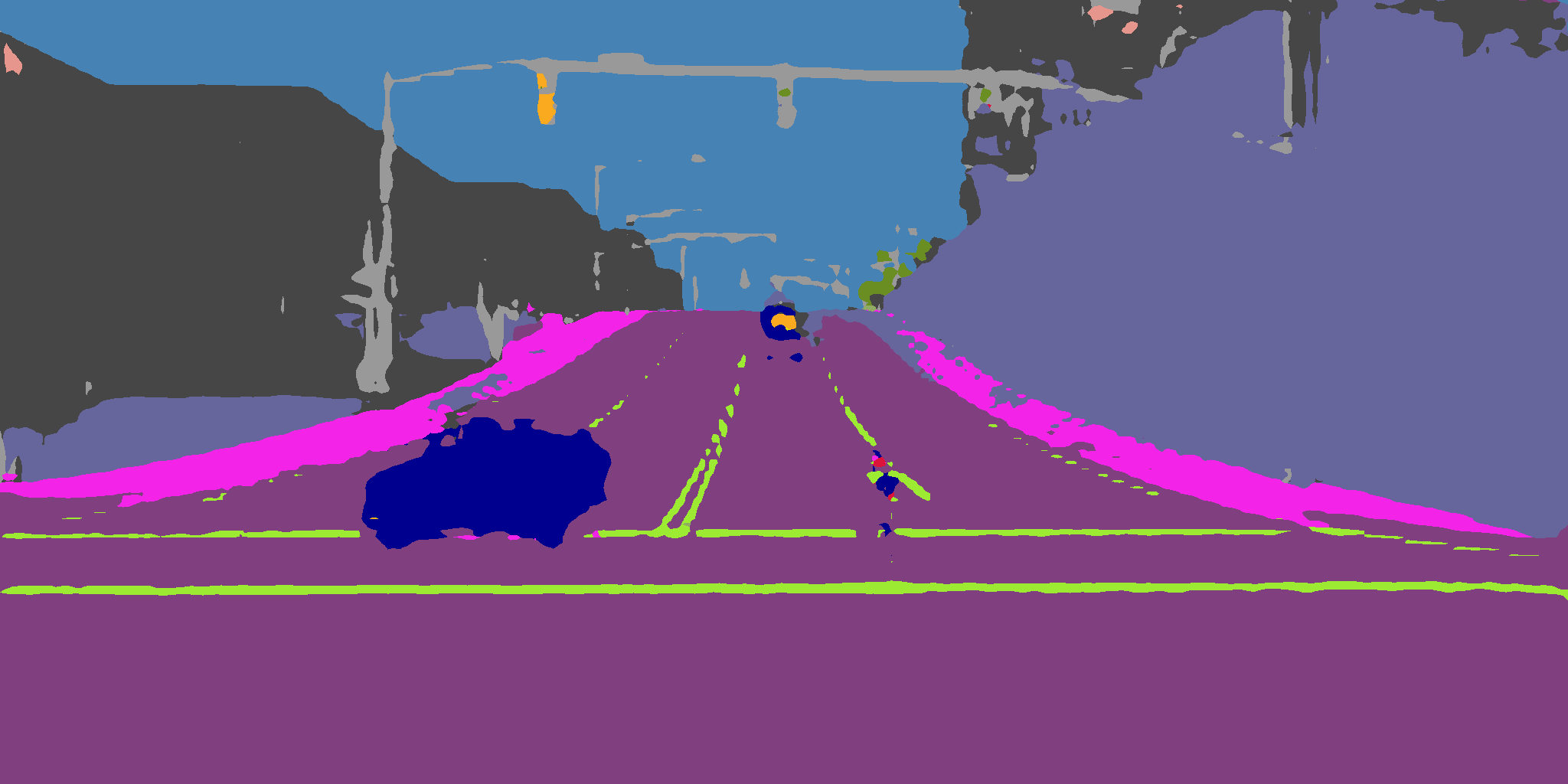}}\\ \vspace{-.6cm}
    \subfloat[][]{\includegraphics[width=0.49\textwidth, trim={0 128px 0 0}, clip]{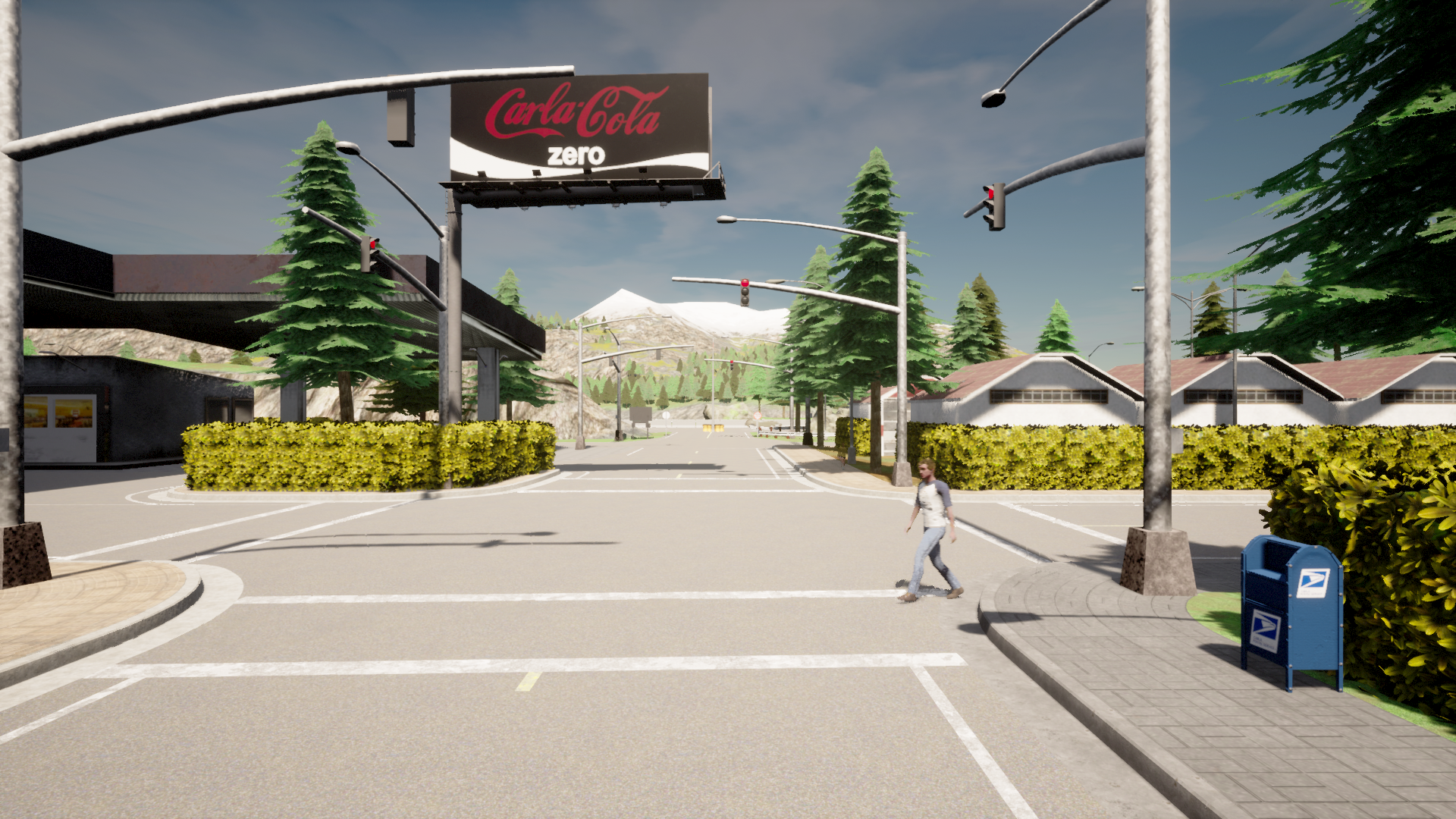}}~~
	\subfloat[][]{\includegraphics[width=0.49\textwidth, trim={0 0 0 0}, clip]{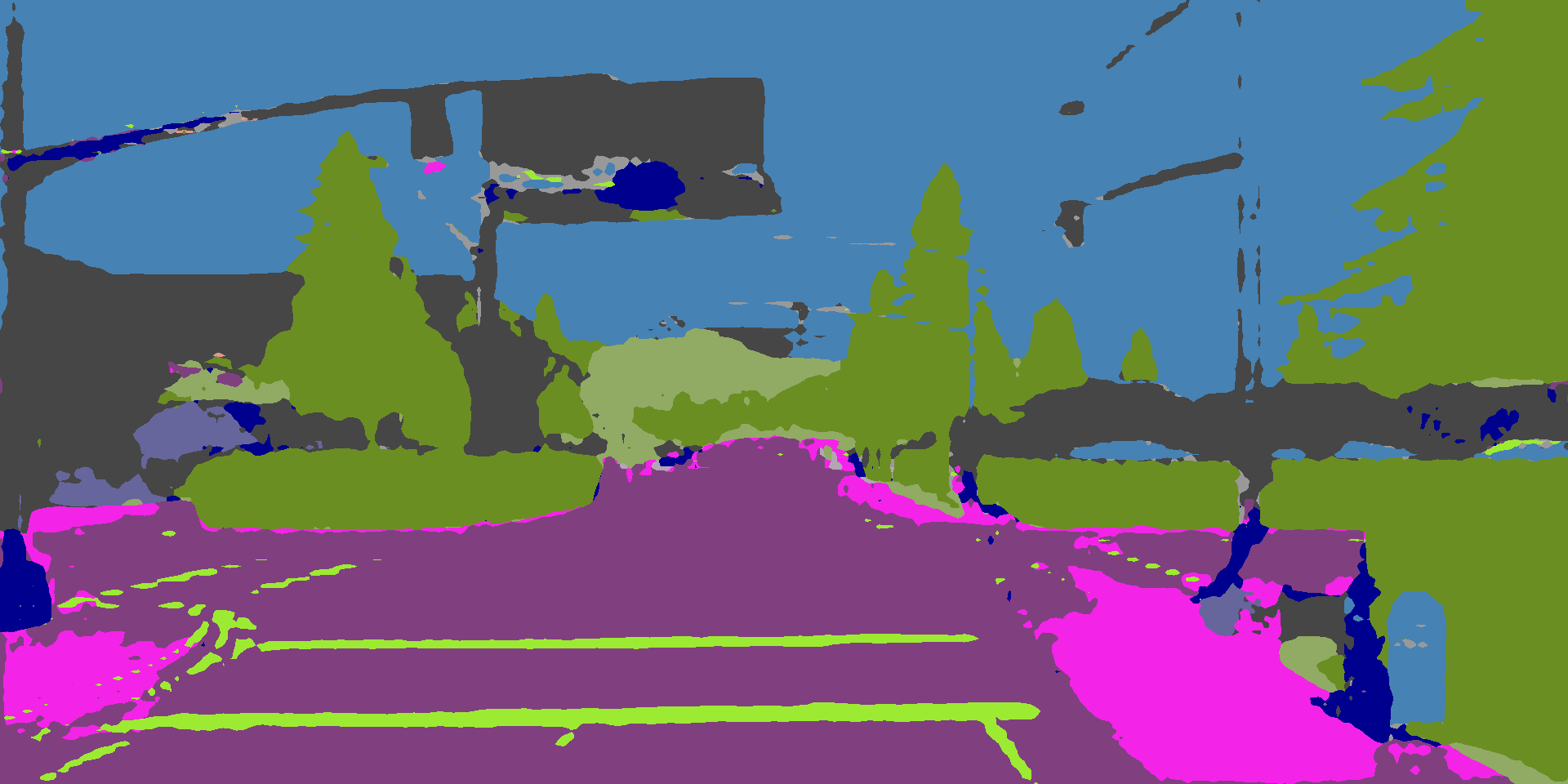}}
	\vspace{-0.6cm}
	\caption[Examples for corner cases]{Two examples of a corner case with pedestrians included, where the safety driver had to intervene to avoid a collision due to the poor prediction of the semantic segmentation network (both images on the right).}
	\label{fig:example_cc} 
\end{figure}

The corner cases were generated as shown in \Cref{fig:flowchart}, using the real-time semantic segmentation network Fast-SCNN where visual perception was limited by intentionally stopping training early. This is sufficient to move in the virtual streets, but is poor enough to enhance corner cases of the \emph{Method Layer}. We note that, according to~\cite{Wang2020}, there were 128 accidents involving autonomous vehicles on the road during test operations in 2014-2018, at least 6\% of which can be directly linked to misbehavior by the autonomous vehicle. It follows that at least every $775335$ km driven, a wrongful behavior of the autonomous vehicle occurs. Using a poorly trained network as a part of our accelerated testing strategy, we were able to generate corner cases after $3.34$ km on average between interventions of the safety driver. We note however that the efficiency of the corner cases was evaluated using a fully trained network. \Cref{fig:example_cc} shows two safety-critical corner cases where the safety driver had to intervene to prevent a collision.
% 1 million miles = 1609344km
% 1609344km*3,7 = 5954572,8
% 5954572,8 / 7,68 (6% of 128 are 7,68)
% = 775335km

In the event of a corner case being triggered by the safety driver, the test operators are required to label the scenario with one of four options (overlooking a pedestrian or a vehicle, disregarding traffic rules, intervening out of boredom) and provide comments. In addition, the duration and the kilometers driven until the corner case appears are registered. The test drivers were instructed to obey traffic rules and not exceed 50 km/h during the test drives. Over time, the drivers became more familiar with the system, leading to a decrease in driving errors and sudden braking. However, a learning effect also occurred where drivers may have hidden situations where objects were not detected by the system. The test rides are tracked and recorded, with the last three seconds of a corner case scenario being saved at 10 fps. This data is then used to retrain the system, with a mix of original and corner case images. 50 corner cases in connection with pedestrians were collected, resulting in 1500 new frames for retraining, with an equal number of frames being removed from the original training dataset.

We were able to show in~\cite{kowol22simulator} that the occurrence of a corner case situation in a model trained with about two-thirds of \emph{Method Layer} corner cases took almost twice as long as in a model trained with the original dataset or with more pedestrians included, see \Cref{tab:2ndcampaign}. The latter was checked because using corner cases with pedestrians results in more pedestrian pixels being available in the data. To allow a fair comparison the additional model was trained with the same average number of pedestrian pixels per scene.

\begin{table}[h!]
	\begin{center}
	\caption[Corner case appearances on Fast-SCNN]{Corner case appearances on Fast-SCNN trained with 3 different datasets. The table was already published in~\cite{kowol22simulator}.}
	\scalebox{0.8}{
		\begin{tabular}{l|c|c|c|c|c|c|c}
			\hline
			\multicolumn{1}{c|}{\multirow{2}{*}{\textbf{dataset}}} & \textbf{distance} & \textbf{time} & \textbf{\#CC} & \textbf{mean}$_{d_{CC}}$ & \textbf{std}$_{d_{CC}}$ & \textbf{mean}$_{t_{CC}}$ & \textbf{std}$_{t_{CC}}$ \\
			\multicolumn{1}{c|}{} & $d$~{[}km{]} & $t$~{[}min{]} & {[}-{]} & {[}km/CC{]} & {[}km/CC{]} & {[}min/CC{]} & {[}min/CC{]} \\ \hline
			\rowcolor[gray]{0.9} natural disritbution & 121.32 & 411 & 13 & 7.73 & 14.25 & 25.93 & 39.60 \\
			pedestrian enriched & 163.09 & 500 & 21 & 7.52 & 10.47 & 23.25 & 28.72 \\
			\rowcolor[gray]{0.9} corner case enriched & 153.38 & 528 & 11 & \textbf{13.84} & 8.68 & \textbf{47.47} & 31.87 \\ 
			\hline
		\end{tabular}
	}
	\label{tab:2ndcampaign}
	\end{center}
\end{table}
We have therefore demonstrated the benefits of our method for generating corner cases, especially for safety-critical situations. We were also able to show that adding safety-critical corner cases recorded by intentional perceptual distortions improves performance, so future datasets should include such situations.
Next, with this test rig setup we investigate whether a single network is required to overcome the so-called domain gap, which describes the difference in data during training and deployment, or whether, for cost and performance reasons, different networks should be used depending on the task. This will be investigated using different weather conditions and survival analysis.

\section{\uppercase{Survival Analysis}}\label{sec:survival}
Survival analysis is the study of lifespans, also survival times, and their influencing factors~\cite{moore2016survival}. It uses statistical methods to investigate time intervals between sequential events. Groups, but also individuals can be considered as the unit of study when an expected event happens during a considered time period like the time from birth until death, the time from entry a clinical trial until death, the time from buying a vehicle until an accident happens, or other use cases. The basic goals of survival analysis are~\cite{Kleinbaum2012}: %S.16
\begin{itemize}
	\item[$\bullet$] estimation and interpretation of survivor or hazard functions
	\item[$\bullet$] comparing survivor and/or hazard functions
	\item[$\bullet$] relationship determination of explanatory variables to lifespans
\end{itemize}

First, some typical terms of survival analysis are introduced with an overview in~\Cref{tab:terms}. 

\begin{table}[h!]
	\begin{center}
    \caption[Terms in Survival Analysis]{Terms in Survival Analysis.}
    \label{tab:terms}
	\scalebox{.8}{%
	\begin{tabular}{l|l}
	\hline
	\textbf{Term}    & \textbf{Explaination} \\ \hline
	\rowcolor[gray]{0.9} observation time & observation period for which start and end points are known \\
	entity & single object or individual of the observed study\\
	\rowcolor[gray]{0.9}event & change in status (e.g. life to death, accident-free to accident) \\ 
	entry & starting state (e.g. birth, date of vehicle purchase) \\  
	\rowcolor[gray]{0.9}failure time T &  exit time of a subject\\ 
	risk set & all test objects in the study  \\ 
	\rowcolor[gray]{0.9} censoring        & \begin{tabular}[c]{@{}l@{}}incomplete information about either entry before or/and \\ event after the observation time\end{tabular} \\ 
	truncation & \begin{tabular}[c]{@{}l@{}}non-observable data that either does not exist or whose entry \\ and exit state have not been observed\end{tabular}\\ 
	\rowcolor[gray]{0.9} lifespan    & duration until an event occurs \\ 
	hazard & \begin{tabular}[c]{@{}l@{}}probability that an observed entity has a certain \\ event at time $t$\end{tabular}\\ 
	% \rowcolor[gray]{0.9} hazard rate & probability that an event will occur at a given time \\ 
	\hline
	\end{tabular}
	}
\end{center}
\end{table}

\begin{figure}[h!]
	\centering
	\scriptsize
	\includegraphics[width=\textwidth]{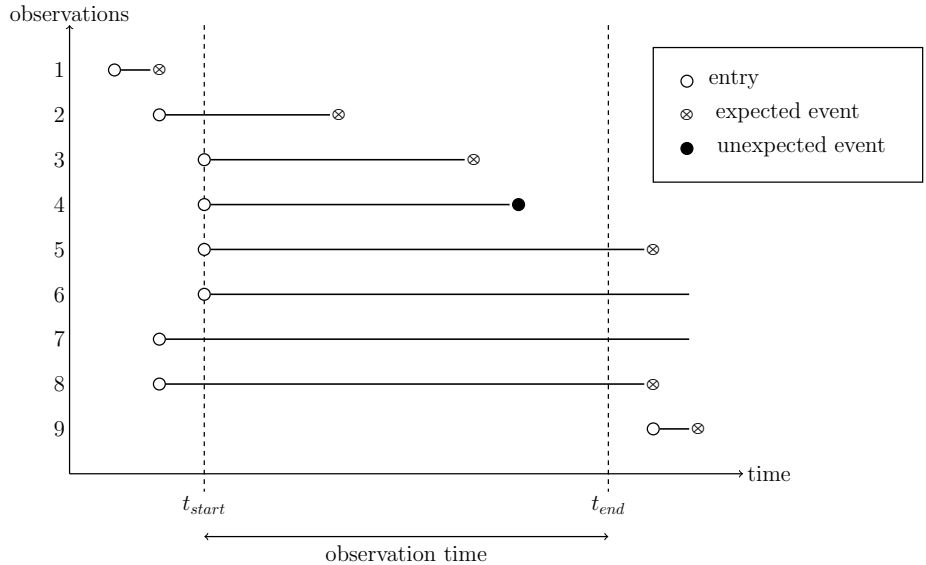}
	\caption[Observation types]{Examples of different observation types. Circles mark the beginning of an observation, while crosses or black circles mark an event. When there is no information about either the entry and/or the event state, this is referred to as censoring.}
	\label{fig:observation_types}
\end{figure}

The observation time period is described by a beginning point $t_{start} = 0$ and an end point $t_{end} > 0 $ defined by a failure condition due to a special event~\cite{hosmer2008survival}. An event implies a change in status, e.g., from alive to dead, from healthy to sick, or from accident-free to accident and is usually easy to find. However, defining the exact failure event is a more difficult task in some cases~\cite{machin2006survival}.
Although it is desirable to know each the beginning and end point of an individual observed in the study, one or both are not always observed which is known as censoring.
\Cref{fig:observation_types} provides an overview of some typical observation types, where white circles describe the entry state. Using our experiments with the driving simulator, the beginning point of pedal pressing may describe the entry state. A cross represents a change of state, such as the occurrence of a corner case due to an impaired perception, while black circles refer to a change of state that was triggered by unexpected reasons like an intervention out of boredom rather than a corner case as cause of impaired perception. 
Observations 1 and 9 describe a \emph{truncated state}, which is non-observable data that either does not exist or whose entry and exit state have not been observed. Observations 2, 7, 8 characterize \emph{left-censored} data as their starting points are not identifiable as they occurred prior the observation start. In addition, observations 5 to 8 escape the observation time unchanged, so they are referred to as \emph{right-censored} as their exit event could not be observed. In addition, the events of observations 2--4 are observed during the observation time, with only 3 being \emph{uncensored} since both start and end times are known. Although an event was detected at observation 4, the expected event did not occur and/or there were other causes for this condition.

Parts of the theory of survival analysis are taken from~\cite{Kleinbaum2012}, unless otherwise stated.
The continuous random variable $T$ describes the time of occurrence of an event, which denotes the time of death of a subject, the time of failure of a machine, start of a disease or similar. $t$ denotes a particular time of interest, which can be used to describe the probability that $T$ has not yet occurred at time $t$, i.e., that the entity has survived. Accordingly, the survival function $S(t)$ represents the probability that the event of an entity at time $t$ did not occur in the observed time period, and can be formulated as follows:
\begin{equation}
	S(t)= Pr(T>t)
	\label{eq:survival_func}
\end{equation}

Two ways to describe a survival distribution are survival and hazard functions.
As a survival function, the so-called Kaplan-Meier~\cite{Kaplan1958} estimator is often used, which estimates the probability that an event for an entity does not occur within a certain time interval. It is defined as follows: 
\begin{equation}
	\hat{S}(t_j) = \prod_{i=0}^{j} \frac{n_i - d_i}{n_i}
	\label{eq:kaplan_meier}
\end{equation}
The observation time $t_j$ is therefore divided into $j$-parts, each of which considers a time interval $\Delta t = (t_i, t_{i+1}]$. With $n$ being denoted by the number of entities which are alive at $\Delta t$ and $d$ the number of entities which already left the observation at $\Delta t$.

Since $T$ is a continuous random variable, it is necessary to work with the probability density function $f(t)$, which describes the probability, that an event occurs in a time interval. The cumulative density function $F(t)$, which is the area under the density function up to the value $t$, describes the probability, that the event occurs at time $T\leq t$:
\begin{equation}
	F(t)= \int_{-\infty}^{t}  f(u)\,du 
	\label{eq:cumulative_func}
\end{equation}
On the other hand, if we consider the probability that an event will not occur until a given time, which is what the survival function means, we can also write the following:
\begin{equation}
	S(t) = 1- F(t)
	\label{eq:surv_fct}
\end{equation}

In many situations, it is crucial to know how an individual risk for a particular outcome changes over time due to other events. For example, weather conditions can negatively affect the lifespan of a semantic driver when the model was not trained with such data. In addition, the use of multiple unknown weather variables can lead to interactions, which in turn can alter a semantic driver's lifespan. For those cases the hazard rate $h(t)$ indicates the probability that an observed entity experiences a failure event the next short time interval $\Delta t$~\cite{Klein2003survivial}. It describes the risk of actual failure rate corresponding as a function over time.

The hazard rate is defined as: 
\begin{equation}
	h(t) = \lim_{\Delta t \to 0+} \frac{Pr(t \leq T< t+\Delta t | t\leq T)}{\Delta t} = \frac{f(t)}{S(t)}
	\label{eq:hazard_func}
\end{equation}

The cumulative hazard $H(t)$ is used to estimate the hazard probability which is defined as follows: 
\begin{equation}
	H(t)= -\log(S(t)) = \int_{0}^{t}  h(s)\,ds 
	\label{eq:cumulative_hazard}
\end{equation}

% Hazard ratios ($HR$) describe the hazard for one individual divided by the hazard for another individual
The Hazard Ratio (HR) is a measure of the relative survival experience of two groups ($A$ or $B$) and is defined as follows: 

\begin{equation}
	HR = \frac{O_A/E_A}{O_B/E_B}
	\label{eq:hazard_ratio}
\end{equation}

The ratio $O/E$ describes the relative death rate of a group, where $O$ is the observed number of deaths and $E$ the number of expected number of deaths. The HR is useful to compare two individuals or groups.

% \paragraph{Cox Proportional Hazards (Cox PH) Model}
The Cox PH model, introduced in 1972~\cite{Cox1972}, uses the hazard function as a function of the influencing variables and looks as follows:
\begin{equation}
	h(t,\textbf{Z}) = h_0(t) \exp(\sum_{i=1}^{p} \beta_i Z_i), \quad \textbf{Z}=(Z_1, Z_2, \dots, Z_p), 
\end{equation}
where $h_0$ describes the baseline hazard, which depends only on time and is therefore equivalent to the Kaplan-Meier estimator. \textbf{Z} denotes the influence variables, which are time-independent and $\beta$ the regression coefficients of the influence variables to be estimated.

The Cox model is often called proportional hazards model since the ratio of the risk for 2 entities with covariates $\mathbf{Z}$ and $\mathbf{Z^*}$ is proportional. The relative risk, also known as the hazard ratio (HR), describes that an individual with risk factor $\mathbf{Z}$ will experience an event proportional to an individual with risk factor $\mathbf{Z^*}$. The relative risk is defined as follows:~\cite{Klein2003survivial}
\begin{align}
	HR = \frac{h(t,\mathbf{Z})}{h(t,\mathbf{Z^*})} 
	&= \frac{h_0(t) \exp(\sum_{i=1}^{p} \beta_i Z_i)}{h_0(t) \exp(\sum_{i=1}^{p} \beta_i Z^*_i)} \\
	&=\exp[\sum_{i=1}^{p} \beta_i (Z_i - Z^*_i)]
\end{align}
It becomes noticeable that HR is independent of time.

Additionally, probabilities about the occurrence of an event can be calculated with the hazard function so that the influence of different parameters can be taken into account. Furthermore, events that have already occurred are included in the calculation so that at a time $d_i$ the probability of an event occurring in the next time step can be predicted. This can be done with the partial likelihood, including a risk set $R(t_d)$ and an index set of death times $D$: 

\begin{align}
	L(\beta) = \prod_{d=1}^{D} \frac{\exp (\sum_{i=1}^{p} \beta_i Z_{di})}{\sum_{j\in R(t_d)} \exp(\sum_{i=1}^{p} \beta_i Z_{ji})}
\end{align}
To optimize the regression coefficients we can maximize the Log-Likelihood: 
\begin{align}
	\beta^* = \argmax \log(L(\beta))
\end{align}

% \noindent 
This is done by computing: 
\begin{align}
	\nabla_{\beta} \log(L(\beta)) = 0
\end{align}
which can be solved numerically.

\subsection{Experimental Design}
\begin{figure}[h!]
    \centering
    \captionsetup[subfigure]{labelformat=empty}
	\subfloat[rain]{\includegraphics[width=0.32\textwidth]  {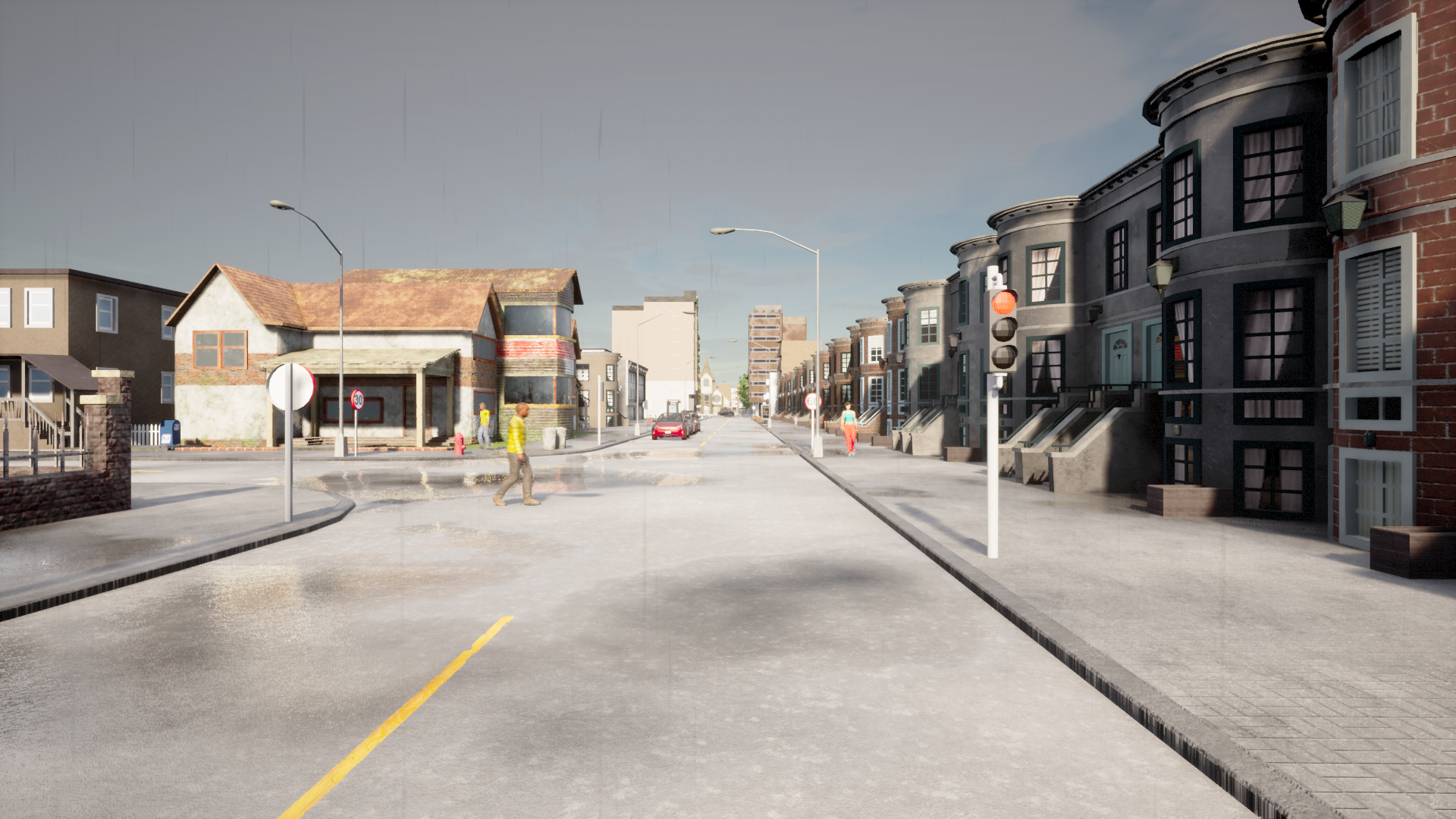}}~~
    \subfloat[fog]{\includegraphics[width=0.32\textwidth]	{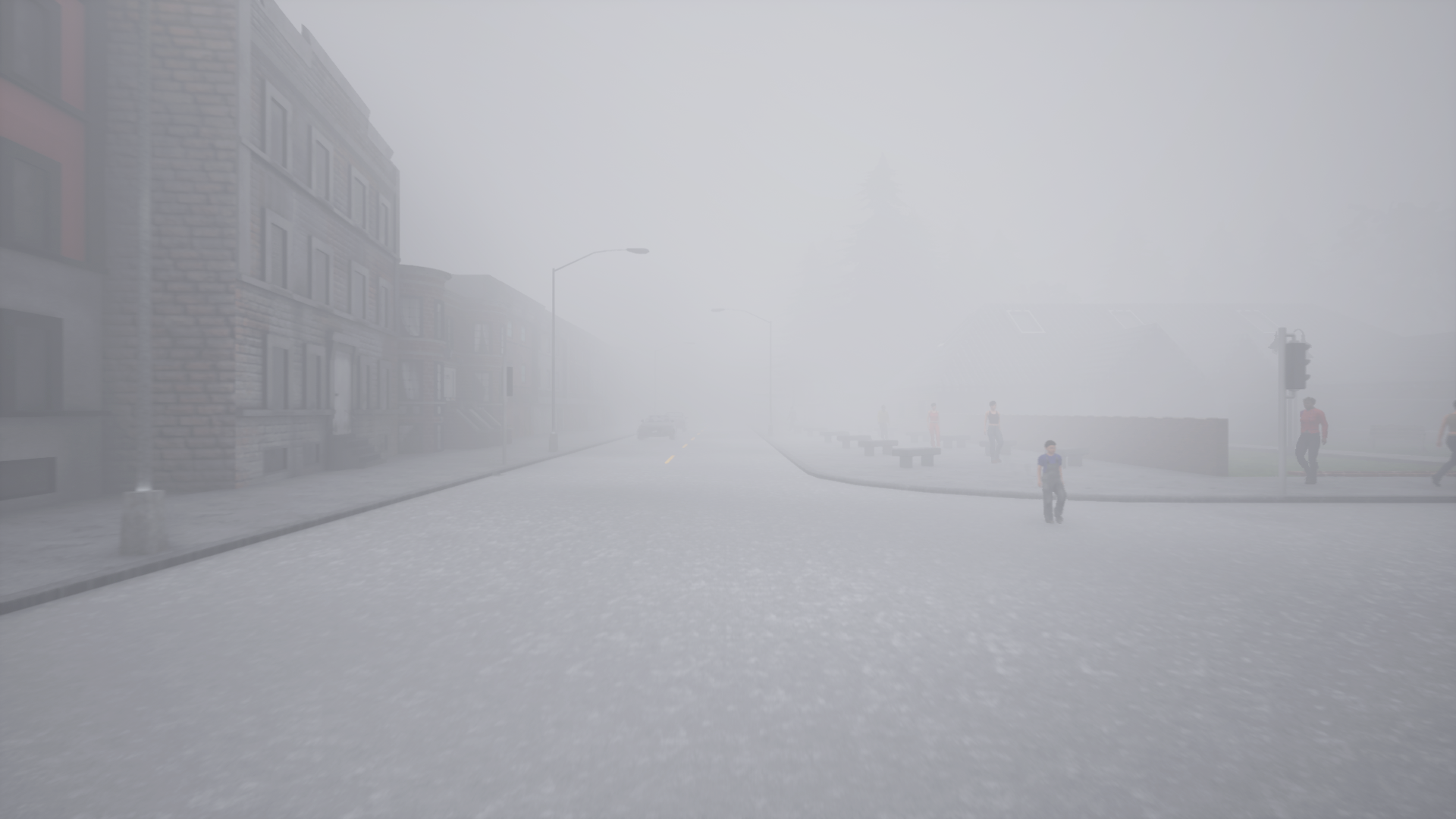}}~~
    \subfloat[night]{\includegraphics[width=0.32\textwidth] {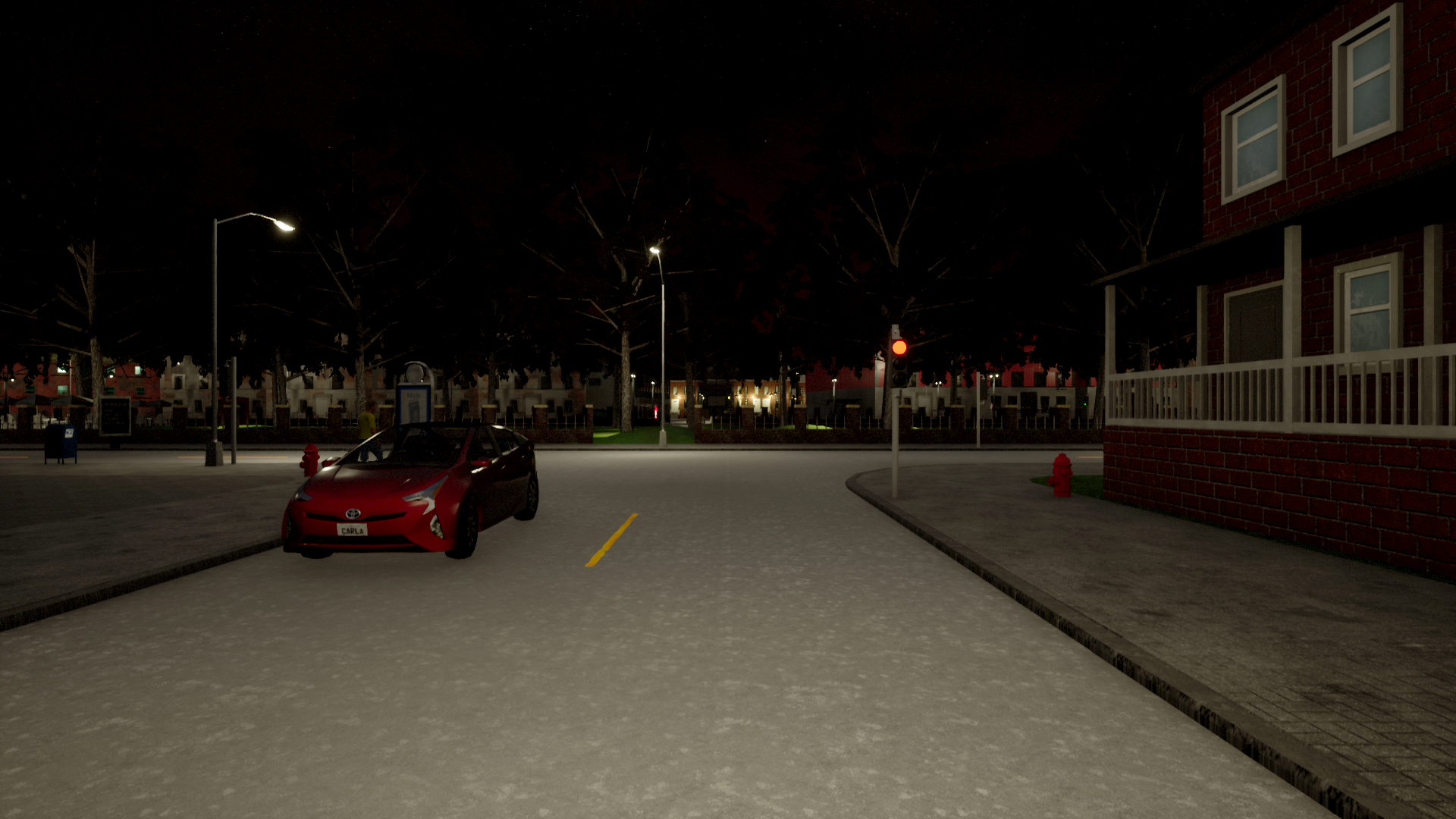}}~~
    \vspace{5pt}
    \caption[Weather conditions for experiments]{Overview of the used weather conditions. The grayish sky, falling water drops as well as water puddles on the road are characteristic for \emph{rain}. In the case of \emph{fog}, fine water droplets cover the image, and it is especially tough to see in depth. \emph{Night} images are characterized by many dark areas, with streetlights and vehicle lights illuminating the scenes.}
    \label{fig:weather}
\end{figure}

After learning the basics of survival analysis, we will use it to find factors that affect survival while driving in the driving simulator. We will use the setup presented in~\Cref{sec:simulator} and observe how long it takes for a corner case to occur under different weather conditions. For this study, the previously used semantic segmentation network Fast-SCNN~\cite{Poudel2019fastSCNN} is trained on good weather data, which we refer to \emph{clear}, and serves as a baseline before being fine-tuned with different weather conditions, namely \emph{rain}, \emph{fog} and \emph{night}.~\Cref{fig:weather} gives an overview of the different weather conditions. In addition, a further model is re-trained on all 3 weather conditions, referred to as \emph{mix}, resulting in a total of 5 models available for the experiments. For post-training, 2100 additional images per weather setting (300 per map) are provided for training and 420 for testing. 
In the following, we refer to each of the weather conditions \emph{rain}, \emph{fog} and \emph{night} as expert models, since they are specifically trained on one domain. In contrast, all 3 weather settings are available to the \emph{mix} model during training, which we refer to universal model. %omniscient model
The baseline and universal models are tested on all five test datasets, whereas the expert models are tested on the respective trained conditions as well as on the \emph{clear} ones. \Cref{tab:performance_testdata} gives an overview of the performance of all models on the particular test data. 
\begin{table}[h!]
	\begin{center}
    \caption{Test data performance for all 5 models.}
	\scalebox{.8}{
	\begin{tabular}{l|p{1cm}p{1cm}|p{1cm}p{1cm}|p{1cm}p{1cm}|p{1cm}p{1cm}|p{1cm}p{1cm}}
		\hline
		model   & \multicolumn{10}{c}{test data} \\  \cline{2-11} 
        & \multicolumn{2}{c|}{\textbf{clear}} & \multicolumn{2}{c|}{\textbf{rain}} &  \multicolumn{2}{c|}{\textbf{fog}} & \multicolumn{2}{c|}{\textbf{night}} & \multicolumn{2}{c}{\textbf{mix}} \\
		& $IoU_{ped}$ & $mIoU$ & $IoU_{ped}$ & $mIoU$& $IoU_{ped}$ & $mIoU$& $IoU_{ped}$ & $mIoU$& $IoU_{ped}$ & $mIoU$ \\
        \hline						%clear              rain fog night mix
		\rowcolor[gray]{0.9} clear 	 & $0.487$ & $0.759$ & $0.368$ & $0.586$ & $0.024$& $0.207$& $0.063$&$0.191$ & $0.123$ & $0.321$ \\
		rain 						 & $0.379$ & $0.606$ & $0.485$&$0.718$ &-& - &-& -&-&- \\
		\rowcolor[gray]{0.9} fog 	 & $0.074$ & $0.130$ &-& -&$0.301$&$0.596$&-&- &-&-\\
		night 						 & $0.292$ & $0.302$ &-& -&-& -&$0.402$&$0.655$&-&-\\
		\rowcolor[gray]{0.9} mix     & $0.451$& $0.657$ &$0.471$ &$0.734$ &$0.326$&$0.644$&$0.369$&$0.694$& $0.402$ & $0.682$ \\
		\hline
	\end{tabular}
	}
	
	\label{tab:performance_testdata}
	\end{center}
\end{table}

The evaluation of the initial model shows a significant decrease of all IoU values in any weather conditions, with the safety-critical class human below $0.1$ for fog and night being awful. 
In contrast, the performance of the universal model remains largely the same. Additionally, compared to the mix model, the expert networks perform better in rain and night and worse in fog for the human class. In the mIoU, the universal model always outperforms the experts. This comparison has already shown the tendency for the expert models to perform at least as well or even slightly better than the universal model in the human class, while the overall performance in the mIoU is best for the universal model in all weather conditions. 
% And this despite the fact that fewer data and time were needed for training.
The next step is to conduct the weather driving campaign, where each model is also tested under these weather conditions in order to obtain a reliable statement about its performance in test.

\begin{figure}[h!]
	\centering
	\centering
	\captionsetup[subfloat]{labelformat=empty}
	\vspace{0.6cm}
	\captionsetup[subfloat]{labelformat=empty}
	\small
	% \subfloat[]{\textbf{clear} \hspace*{2.5cm} \textbf{rain} \hspace*{2.5cm} \textbf{fog} \hspace*{2.5cm} \textbf{night}}\\
	\subfloat[\textbf{clear}]{\includegraphics[width=.24\textwidth]{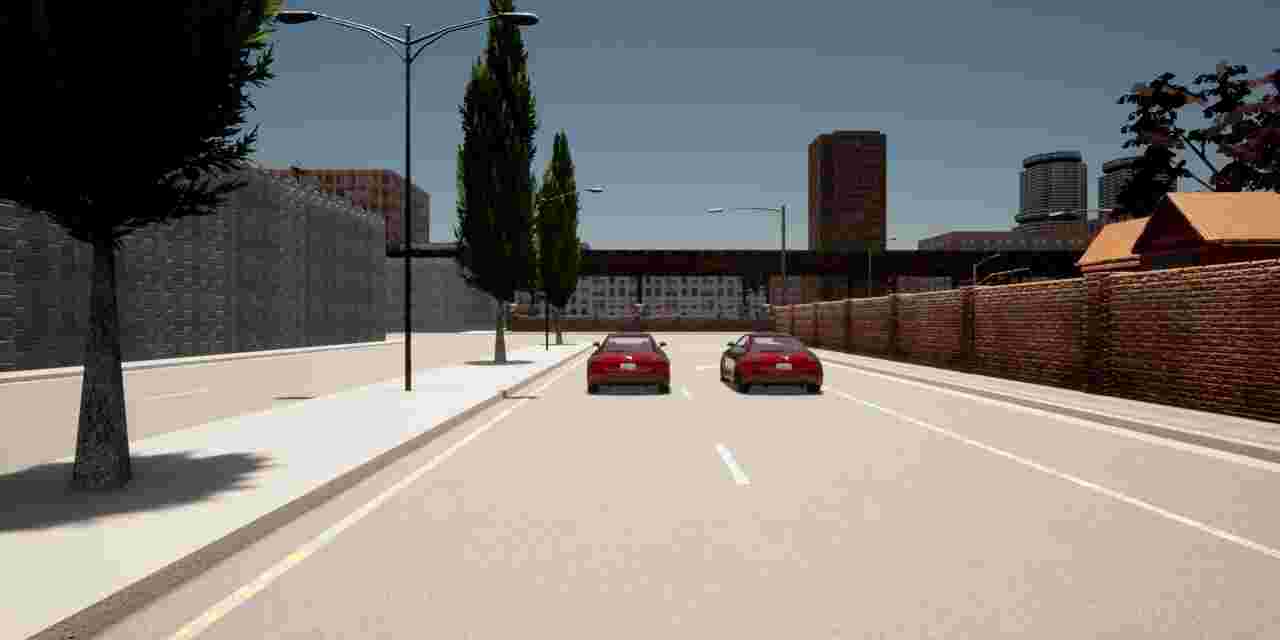}} \hspace{1pt}
	\subfloat[\textbf{rain}]{\includegraphics[width=.24\textwidth]{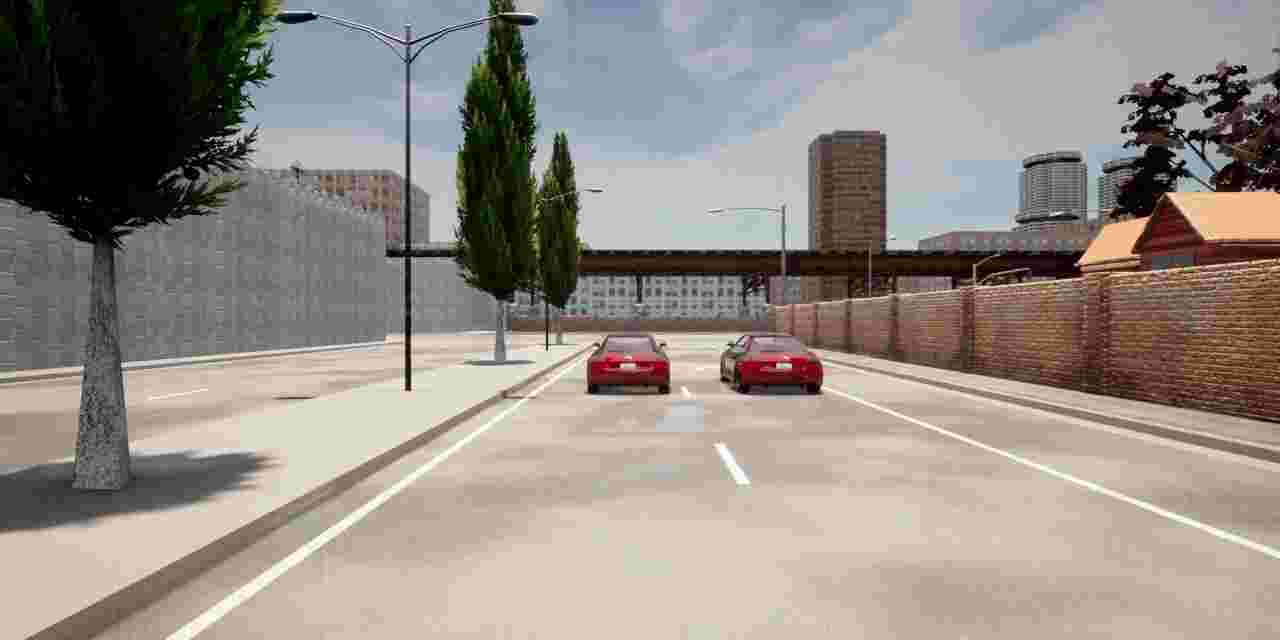}} \vspace{-0.6cm}
	\subfloat[\textbf{fog}]{\includegraphics[width=.24\textwidth]{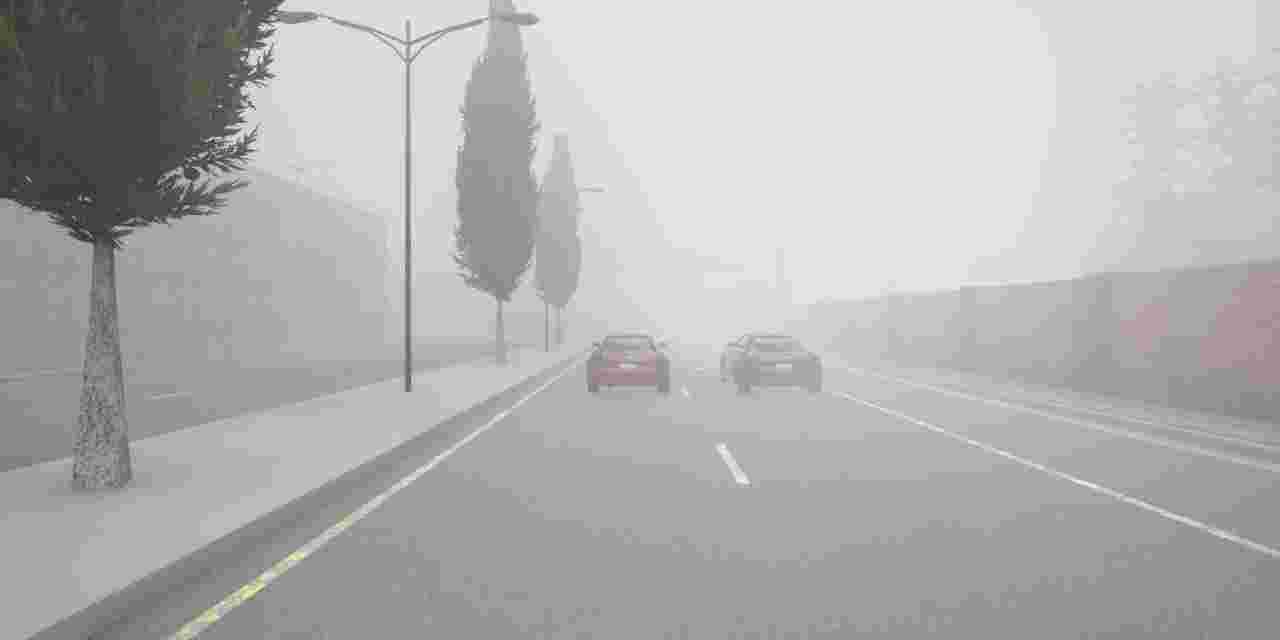}} \hspace{1pt}
	\subfloat[\textbf{night}]{\includegraphics[width=.24\textwidth]{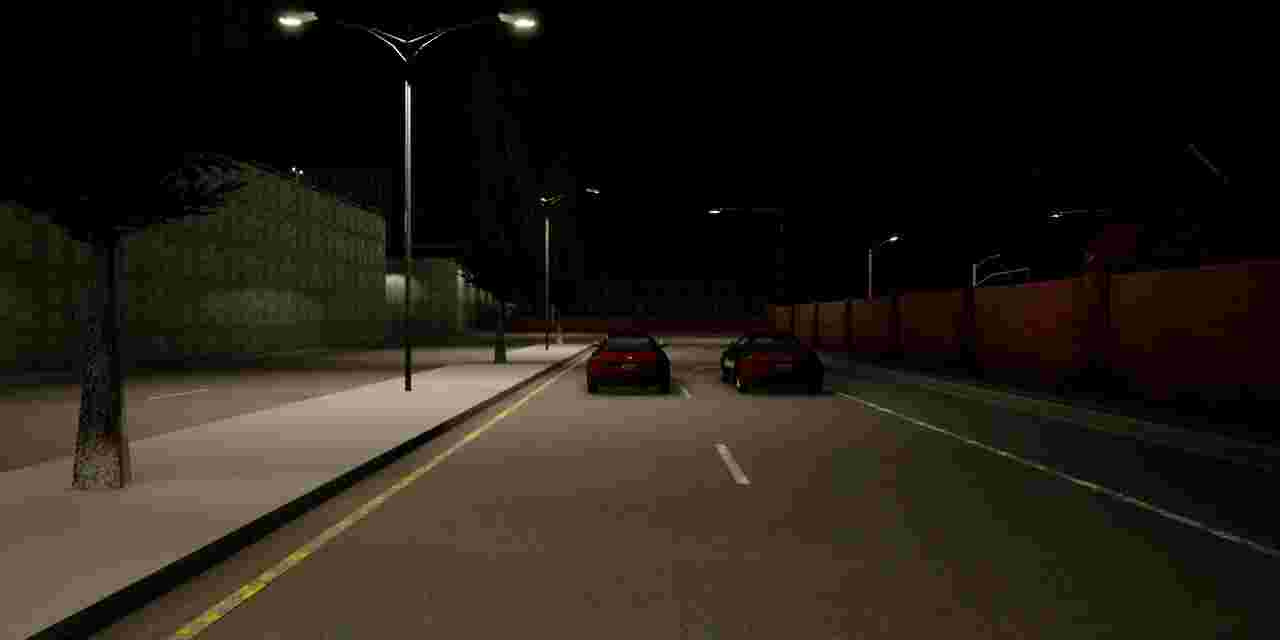}} 
	\vspace{0.6cm} \\

	% \subfloat[]{\rotatebox[origin=lb]{90}{~~clear}}~~
	\subfloat[]{\includegraphics[width=.24\textwidth]{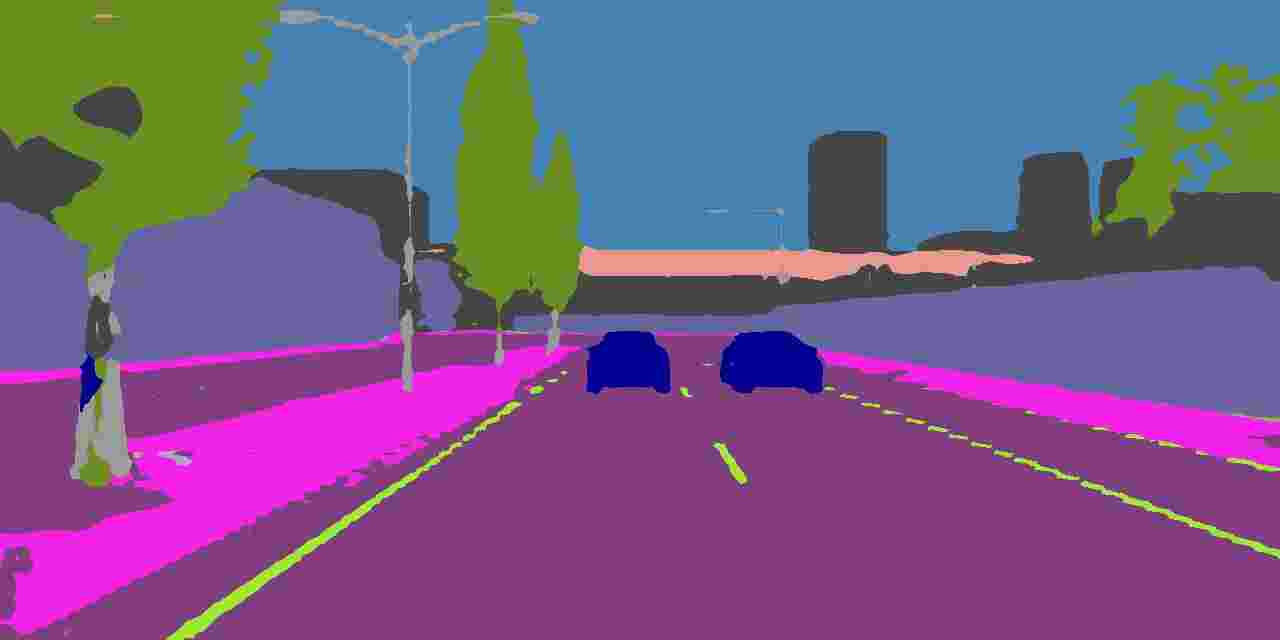}} \hspace{1pt}
	\subfloat[]{\includegraphics[width=.24\textwidth]{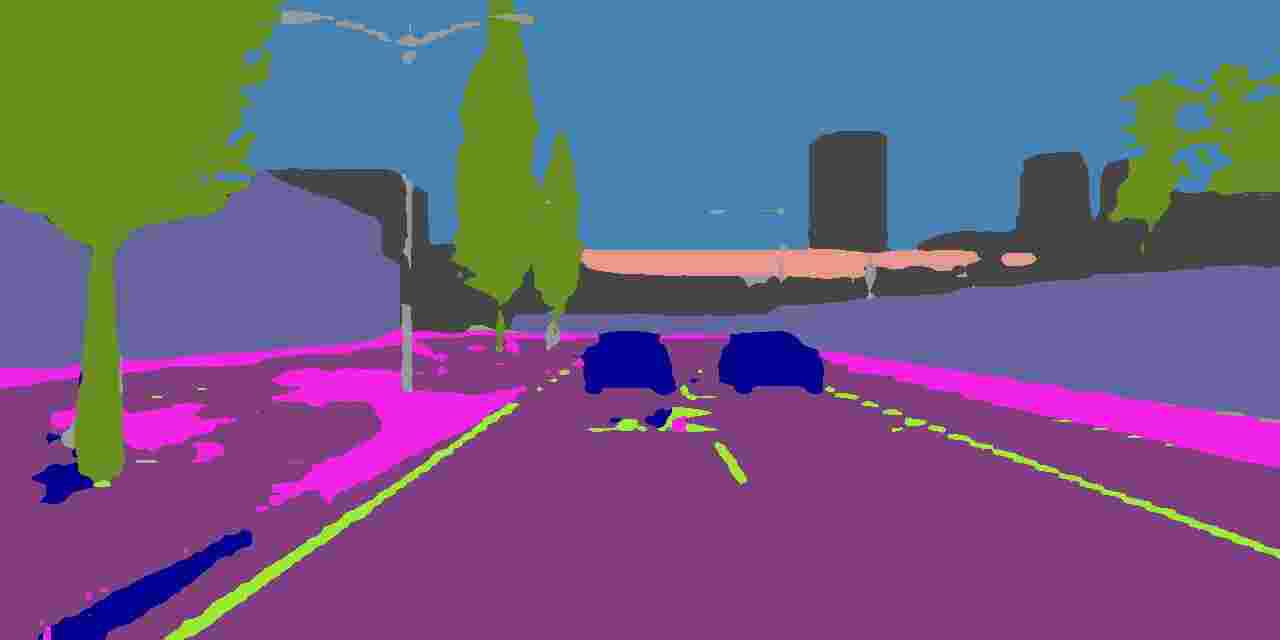}} \vspace{-0.6cm}
	\subfloat[]{\includegraphics[width=.24\textwidth]{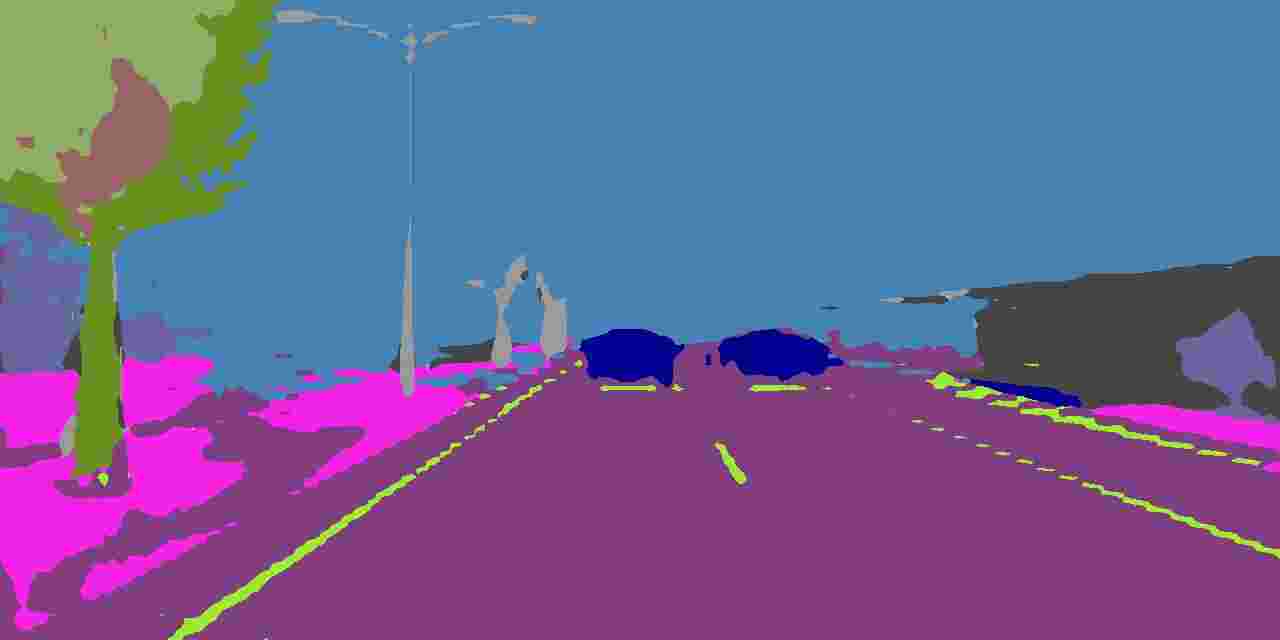}} \hspace{1pt}
	\subfloat[]{\includegraphics[width=.24\textwidth]{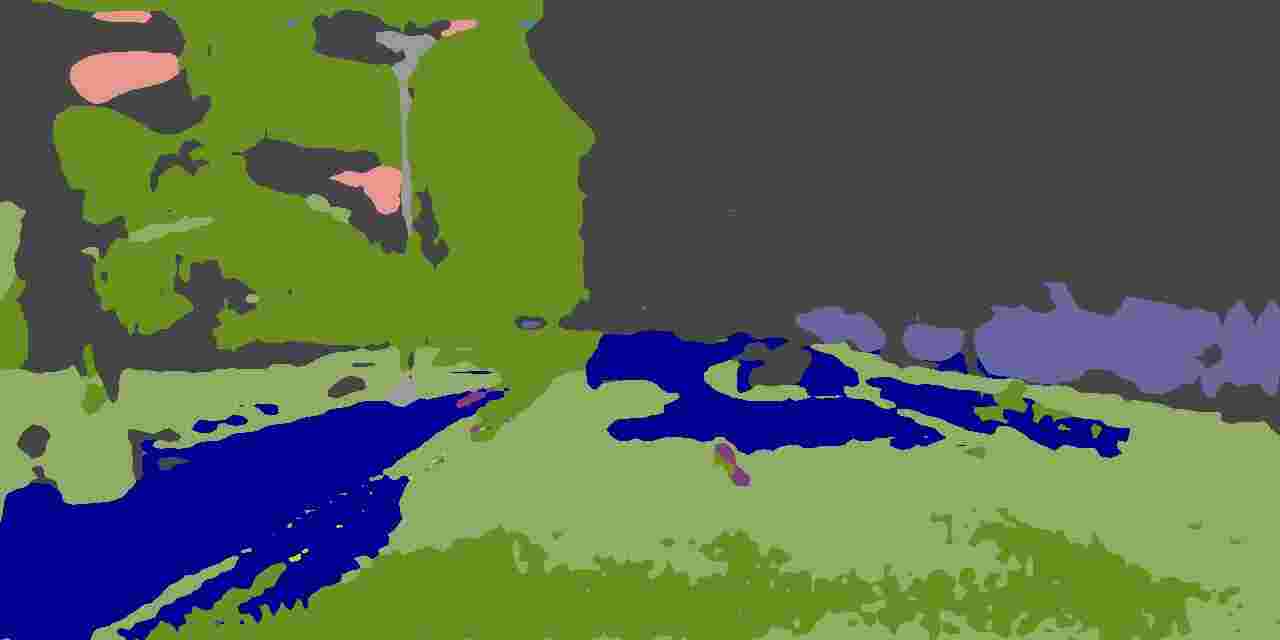}} 
	\vspace{0.2cm}\\  baseline model

	\subfloat[]{\includegraphics[width=.24\textwidth]{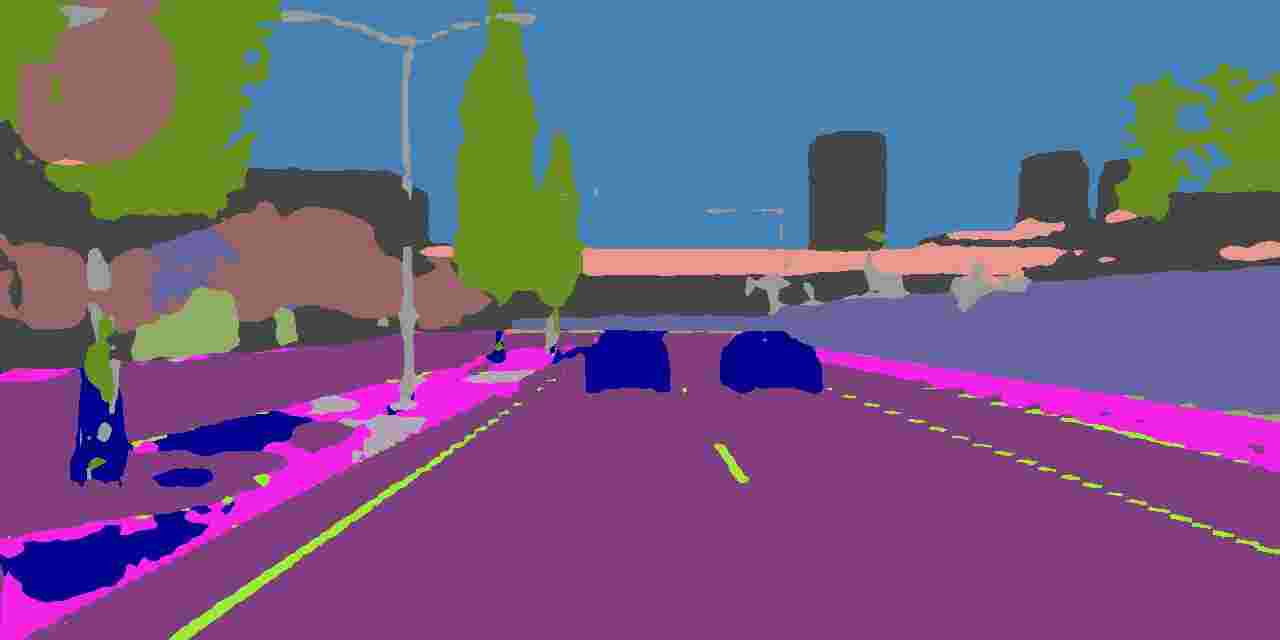}} \hspace{1pt}
	\subfloat[]{\includegraphics[width=.24\textwidth]{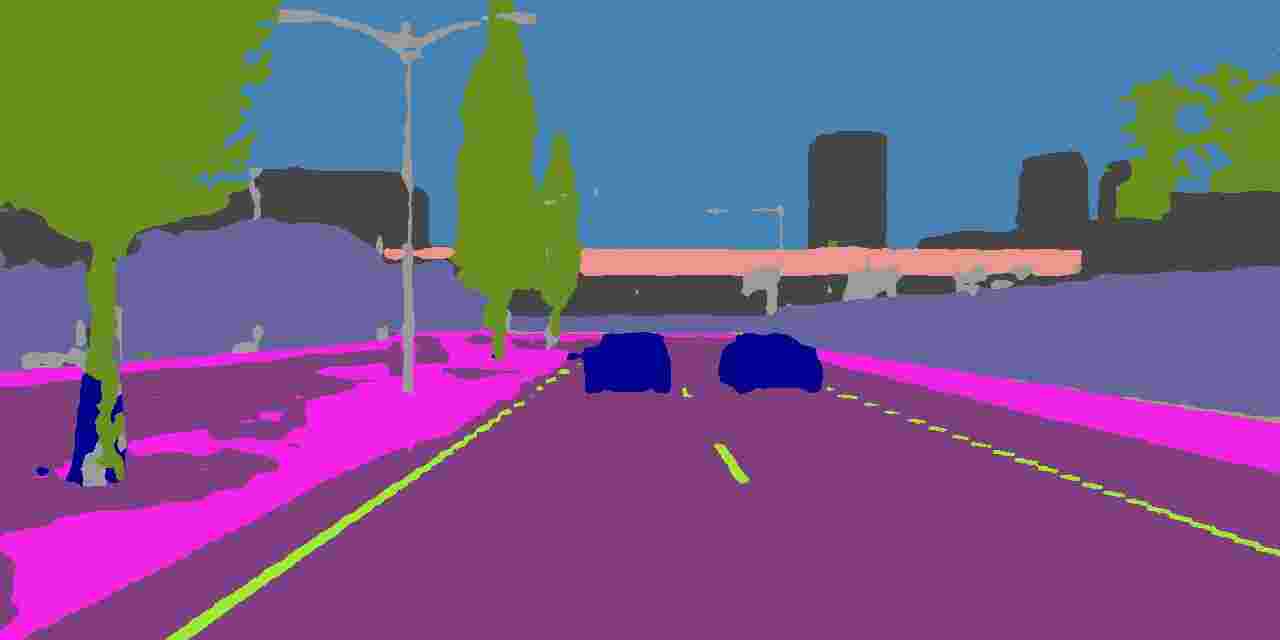}} \vspace{-0.6cm}
	\subfloat[]{\includegraphics[width=.24\textwidth]{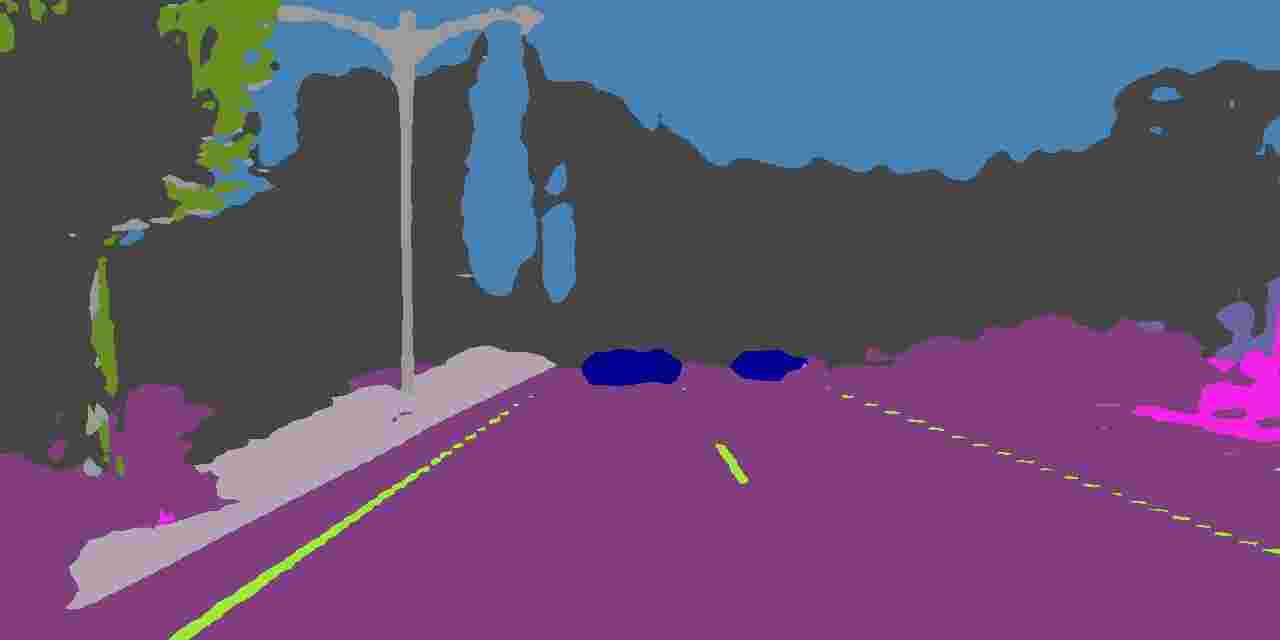}} \hspace{1pt}
	\subfloat[]{\includegraphics[width=.24\textwidth]{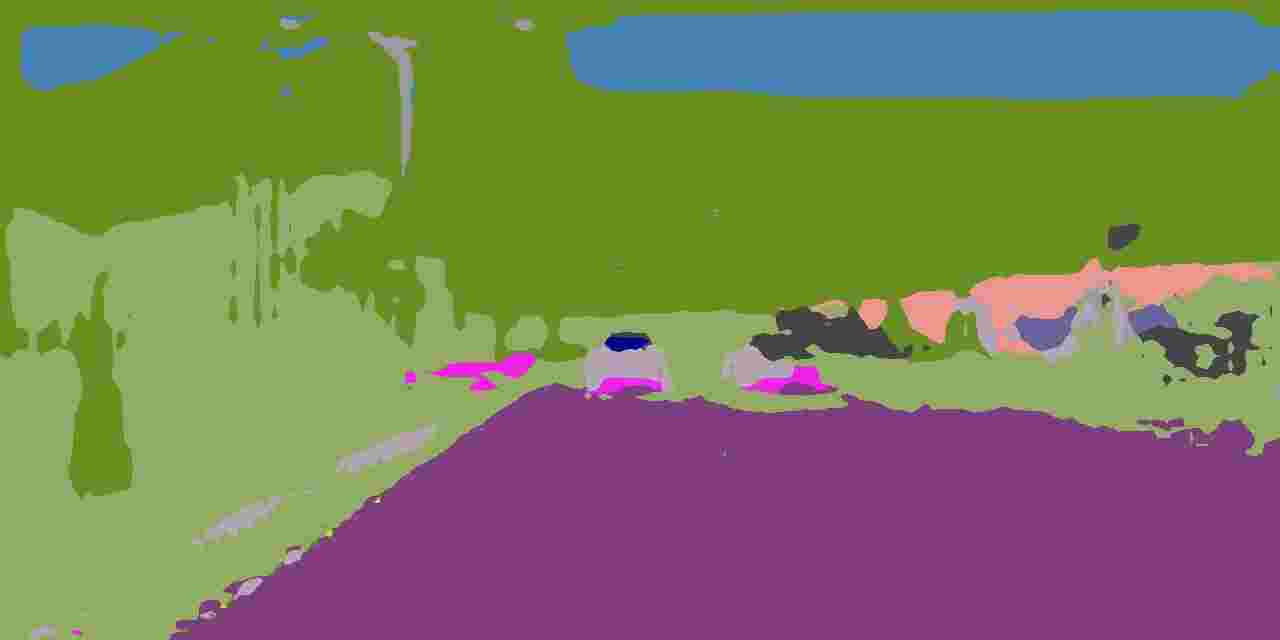}} 
	\vspace{0.2cm}\\ expert model rain
	
	\subfloat[]{\includegraphics[width=.24\textwidth]{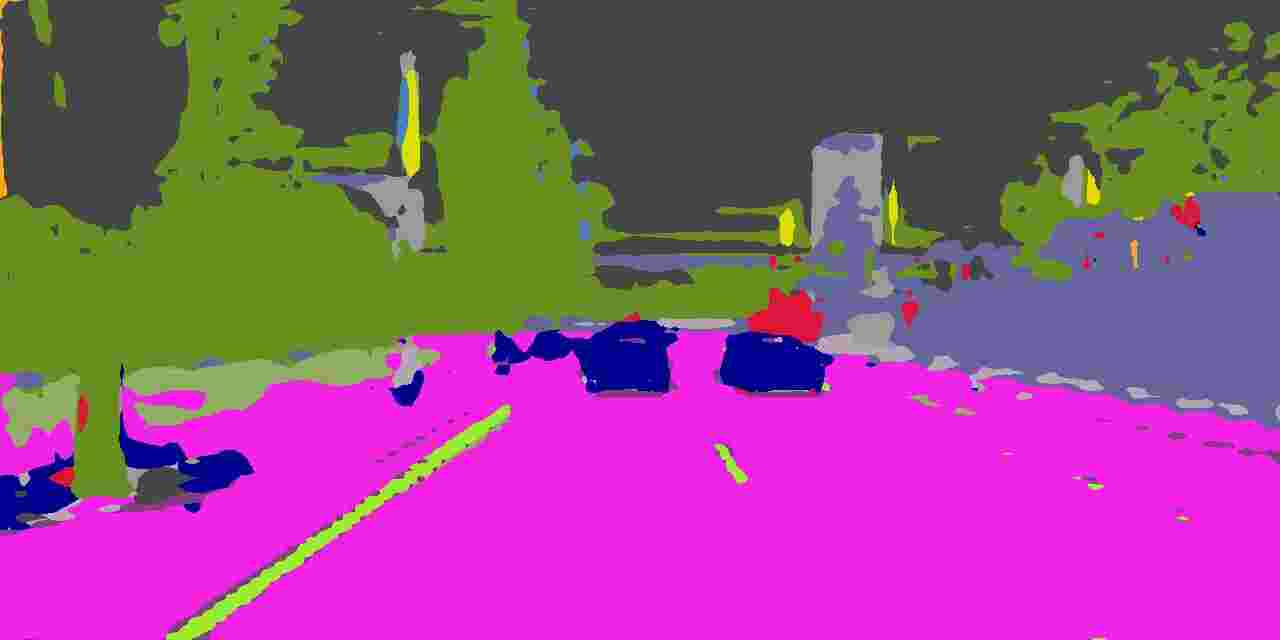}} \hspace{1pt}
	\subfloat[]{\includegraphics[width=.24\textwidth]{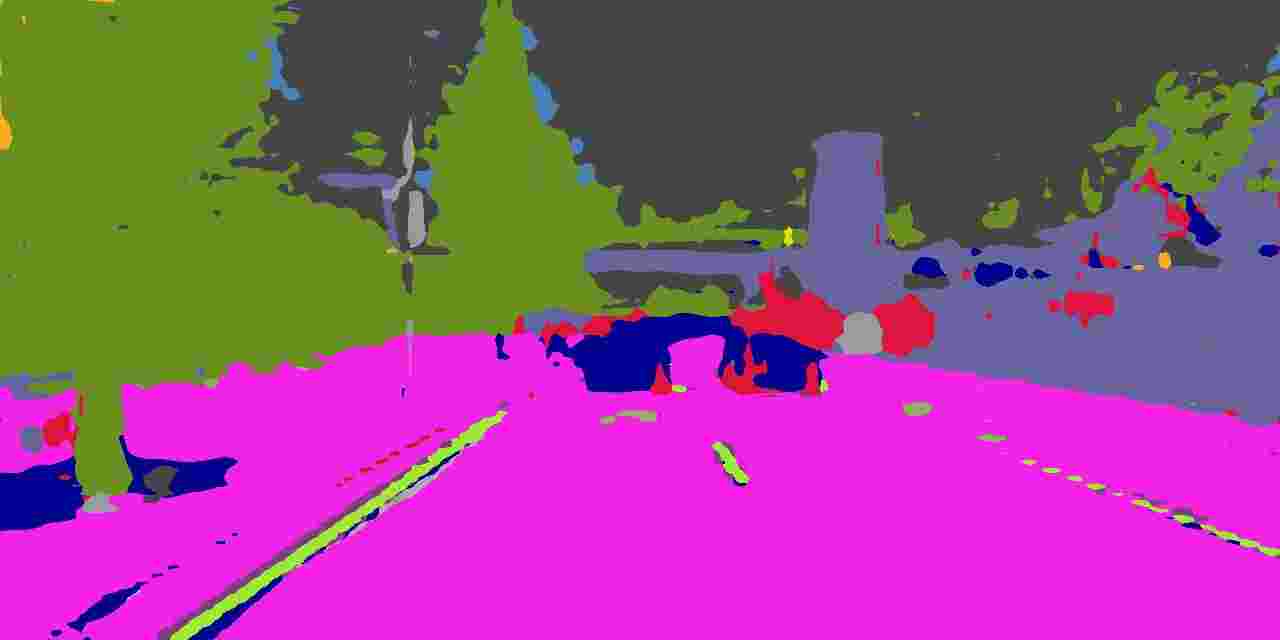}} \vspace{-0.6cm}
	\subfloat[]{\includegraphics[width=.24\textwidth]{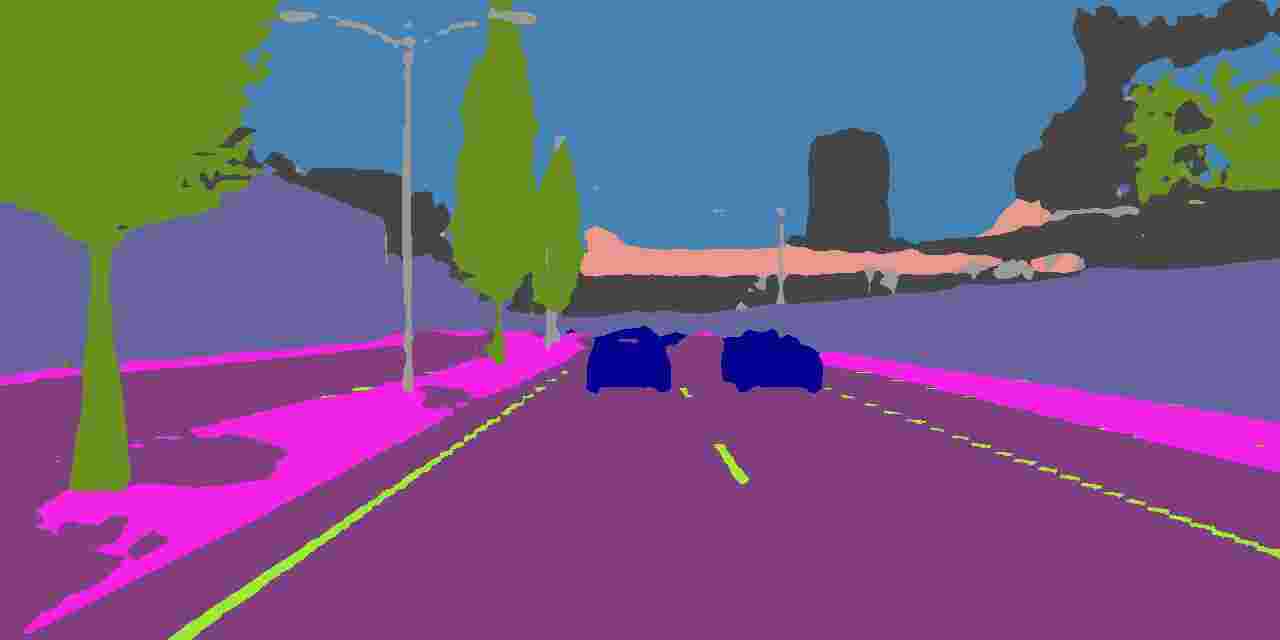}} \hspace{1pt}
	\subfloat[]{\includegraphics[width=.24\textwidth]{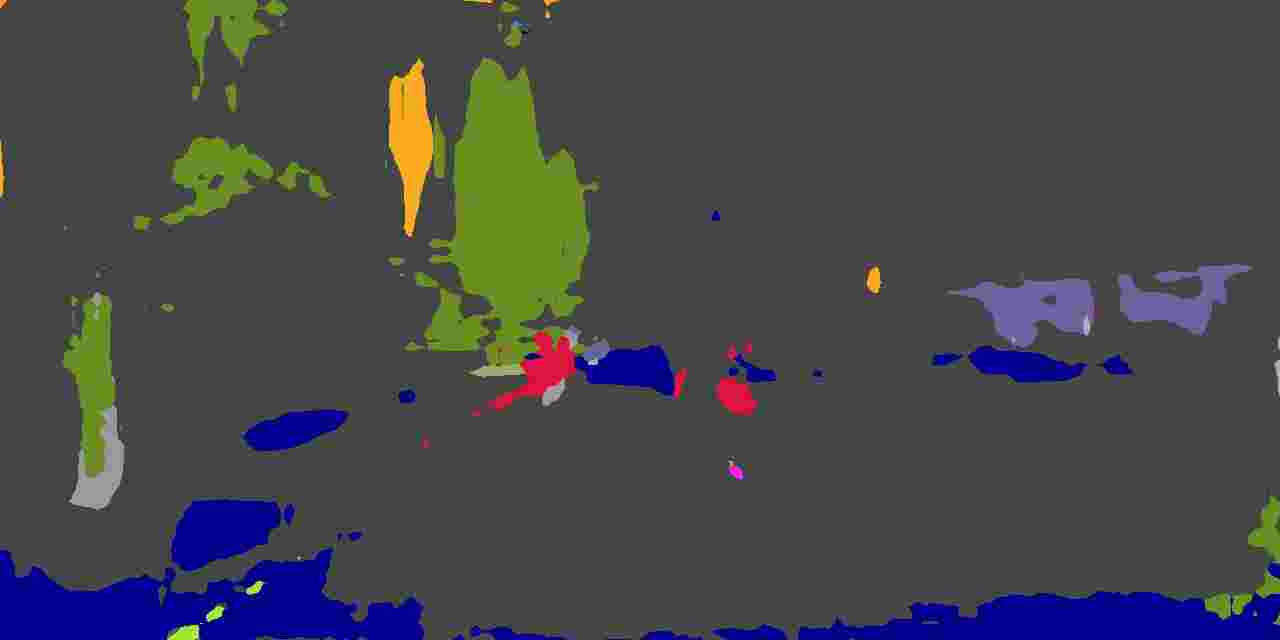}} 
	\vspace{0.2cm}\\ expert model fog
	
	\subfloat[]{\includegraphics[width=.24\textwidth]{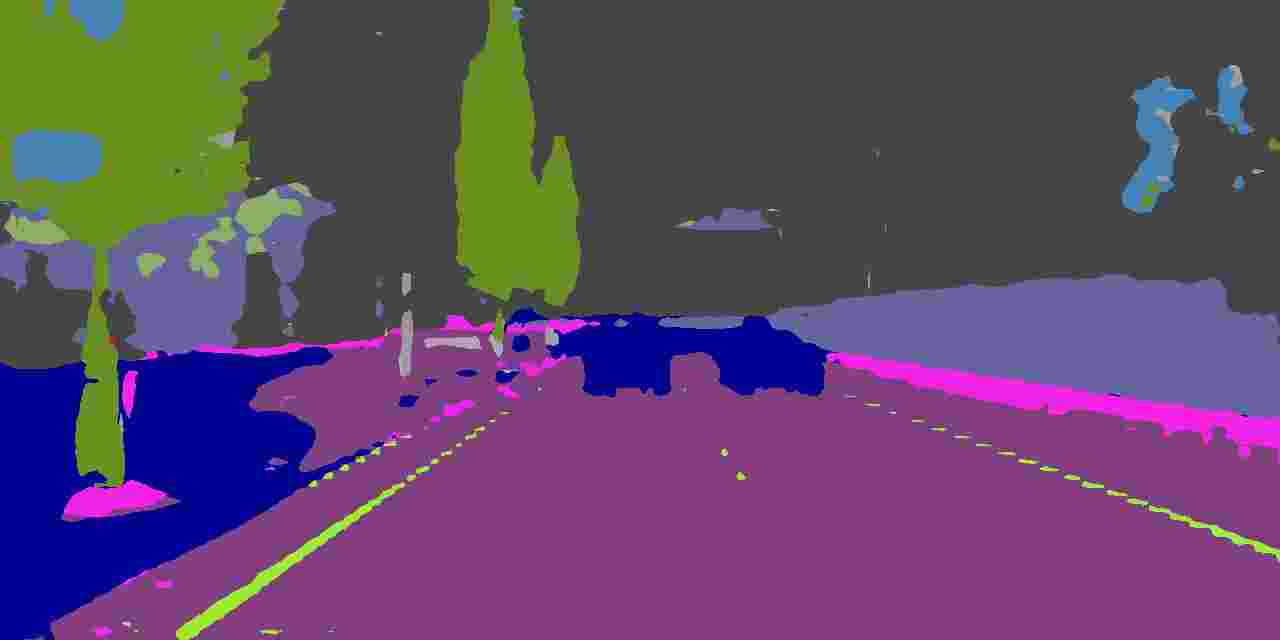}} \hspace{1pt}
	\subfloat[]{\includegraphics[width=.24\textwidth]{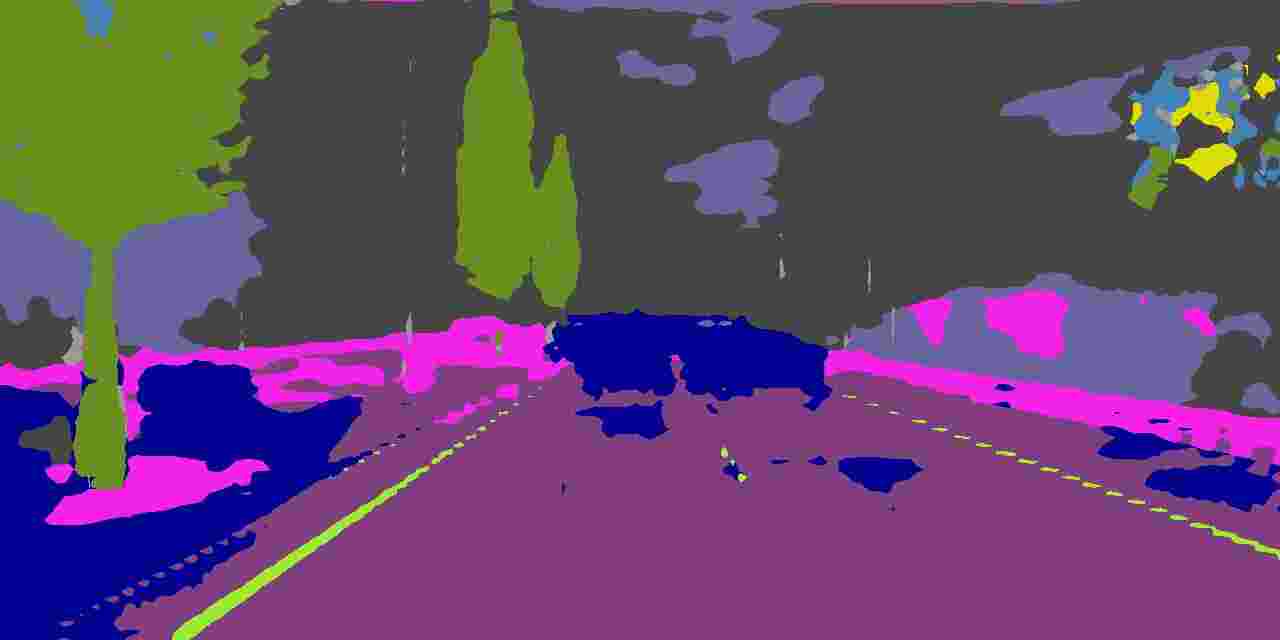}} \vspace{-0.6cm}
	\subfloat[]{\includegraphics[width=.24\textwidth]{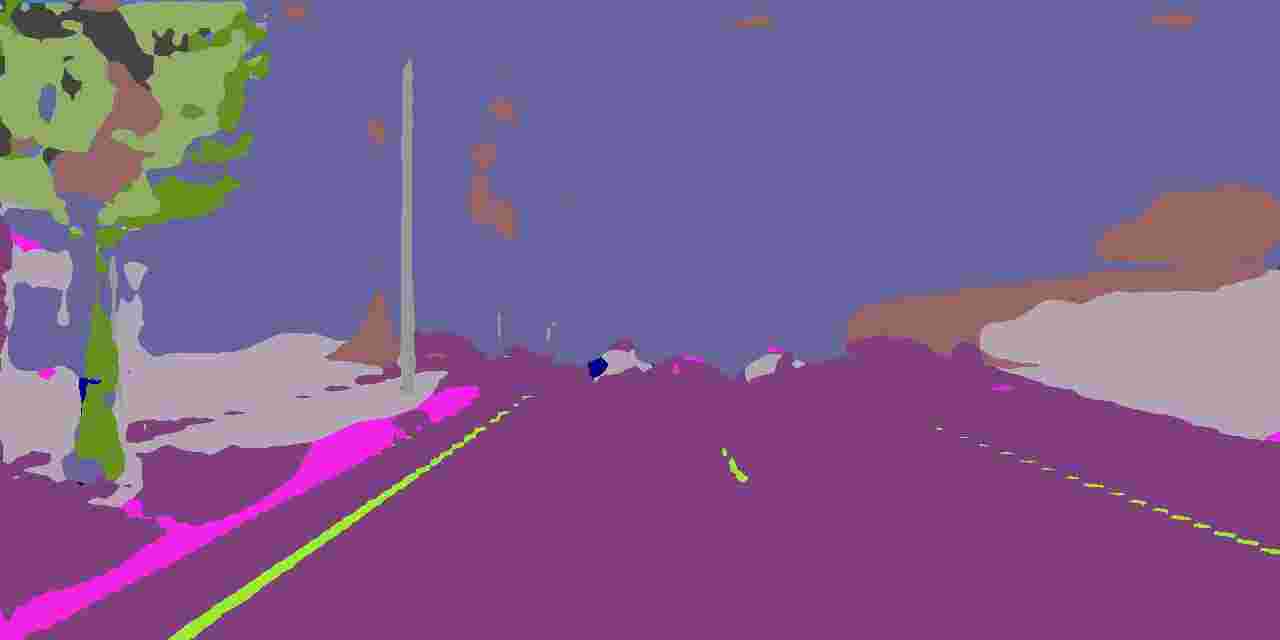}} \hspace{1pt}
	\subfloat[]{\includegraphics[width=.24\textwidth]{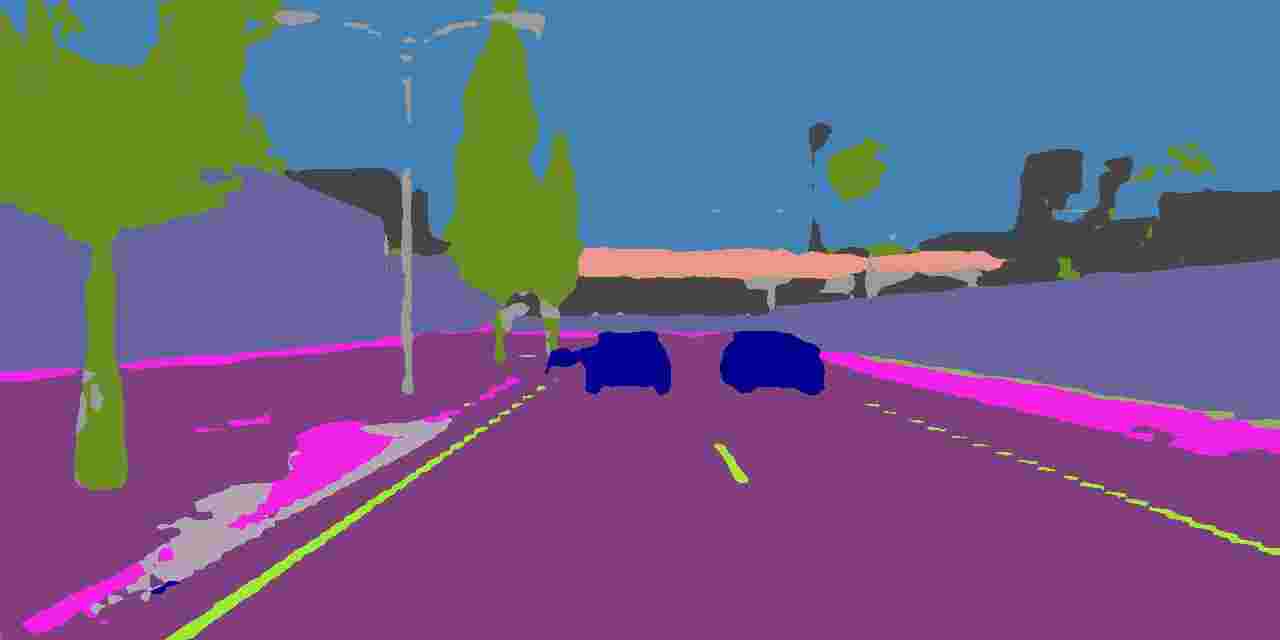}} 
	\vspace{0.2cm}\\ expert model night
	
	\subfloat[]{\includegraphics[width=.24\textwidth]{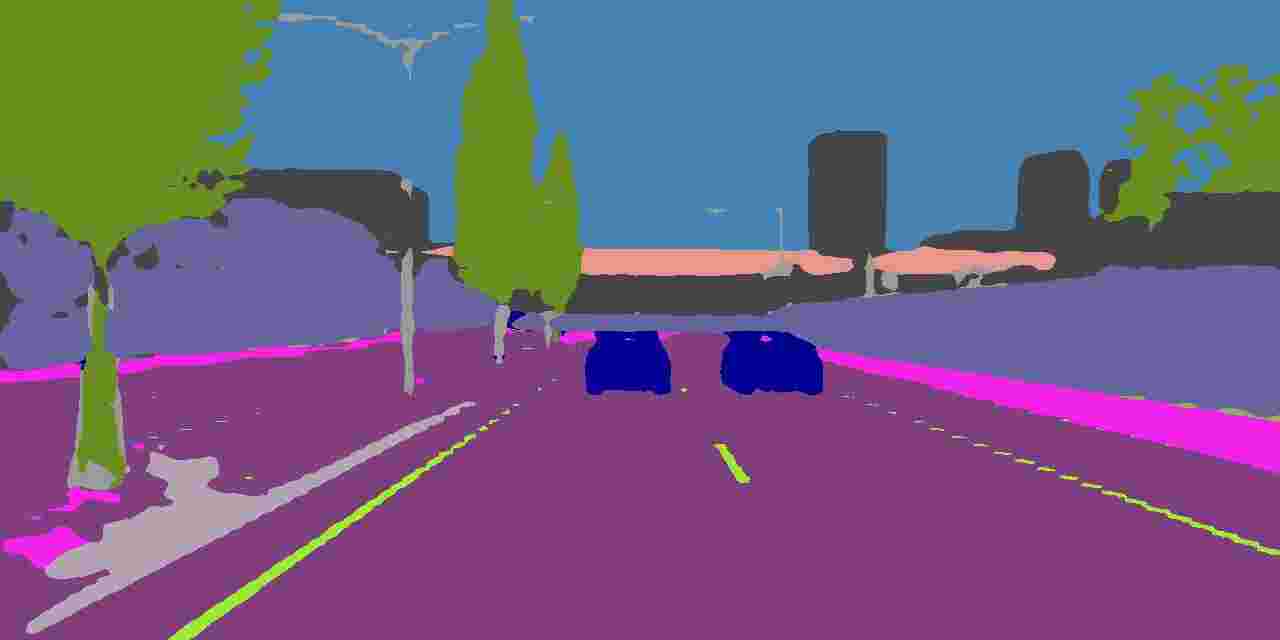}} \hspace{1pt}
	\subfloat[]{\includegraphics[width=.24\textwidth]{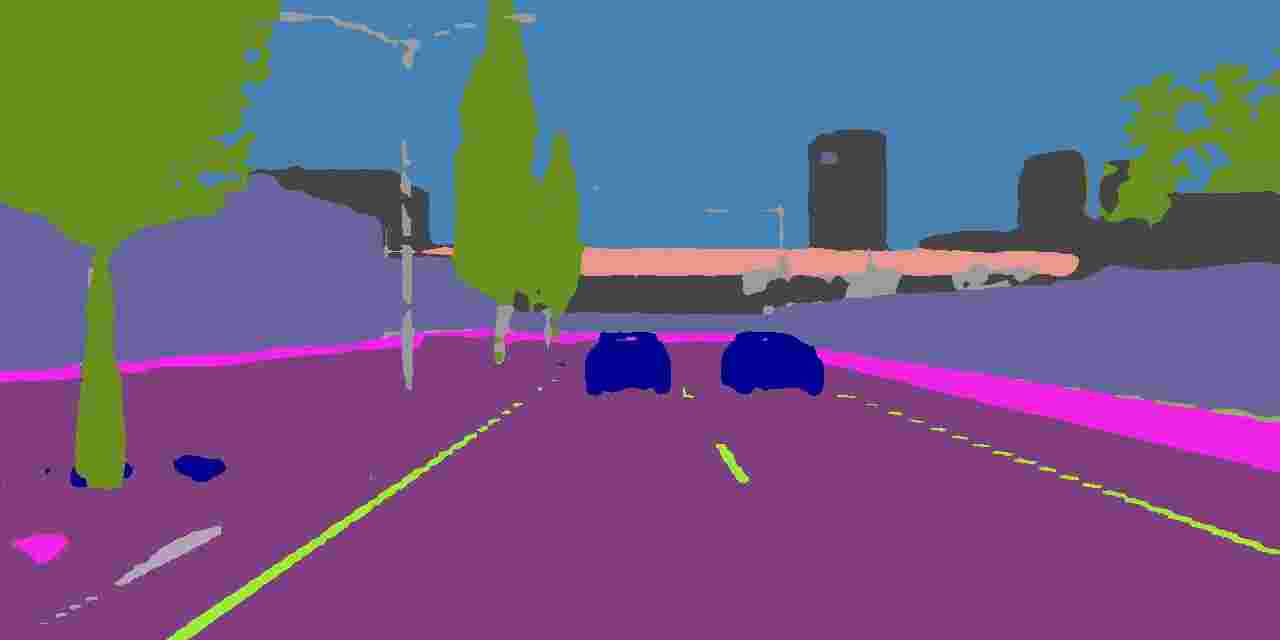}} \vspace{-0.6cm}
	\subfloat[]{\includegraphics[width=.24\textwidth]{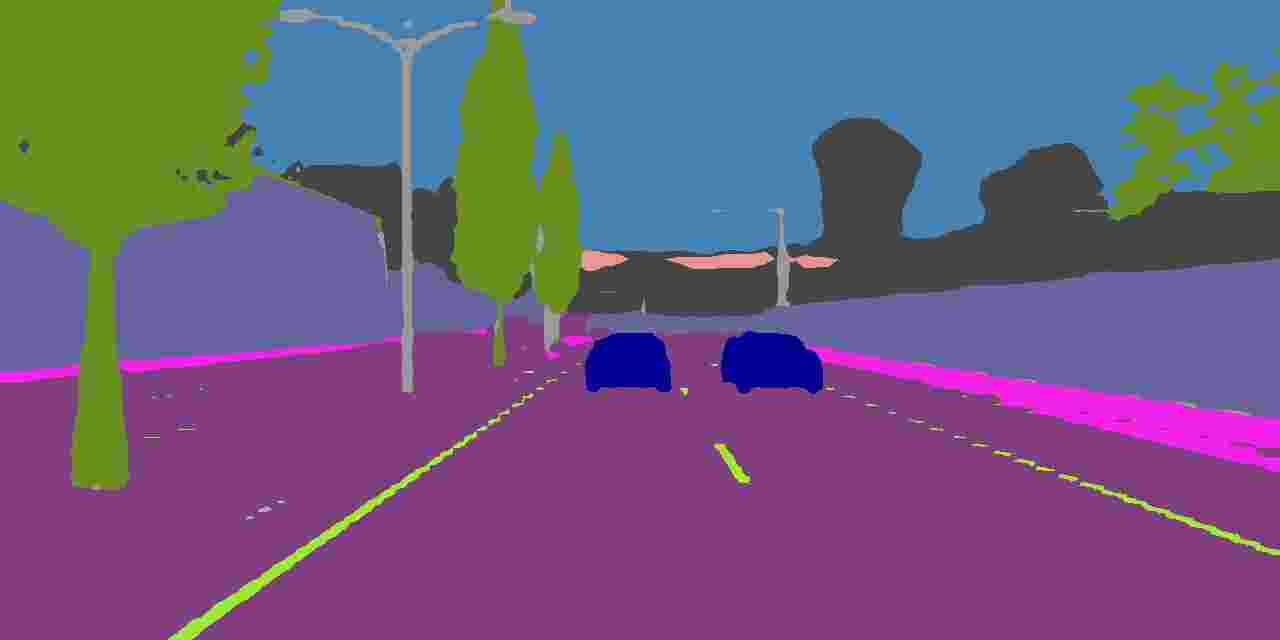}} \hspace{1pt}
	\subfloat[]{\includegraphics[width=.24\textwidth]{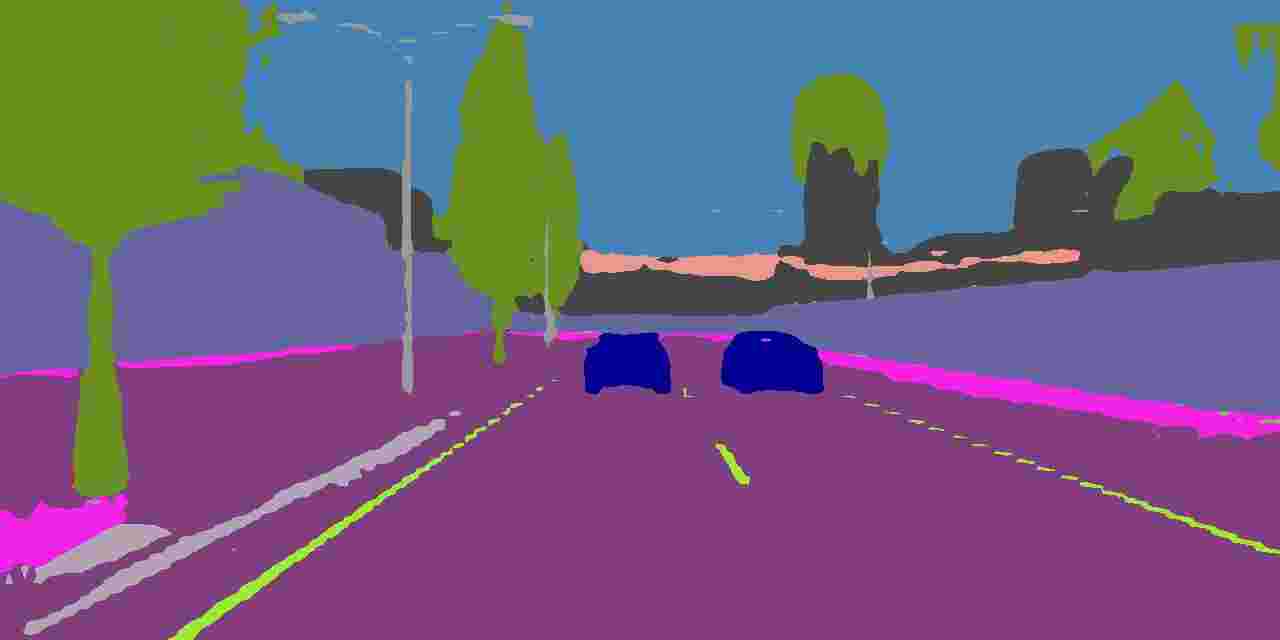}} 
	\vspace{0.2cm}\\ universal model
	
	\caption[Overiew of models and weather]{Model outputs on each weather setup. Under the baseline model, it would be still possible to drive in rain, whereas fog and night would become a risk. The expert models perform well in their domain but quite worse in the other ones. On the other hand, the universal model performs sufficiently well in all weather conditions. }
	\label{fig:overview_models_weather}
\end{figure}

The experiments are conducted as described in~\cite{kowol22simulator}, so that two drivers drive freely on the roads of CARLA. During the rides, the semantic driver has full control over the vehicle, while the safety driver observes the rides and should intervene in the scene only in safety-critical driving situations using the brake pedal or the steering wheel. Intervention indicates incorrect assessment of the scene, which is a corner case of the \emph{Method Layer}. Differences from the previous driving campaigns include the number of maps and the duration of the rides.
This time, the focus is only on Town01 and Town03, since they have a high variability and due to their moderate size the number of vehicles and pedestrians does not need to be set excessively high in order to consistently see some, which relieves the traffic manager and thus computations on the CPU. In addition to the reduced number of maps, the drives will be limited to 600 seconds. If no corner case occurs during this time, the drive is stopped, which corresponds to a right-censored observation. In addition, the drivers didn't know what data the network had been trained on during the experiments as well as what weather condition they were driving in.
The baseline and universal models are tested for 120 minutes on each weather setting (\emph{clear}, \emph{rain}, \emph{fog}, \emph{night}). In addition, the expert models are tested on the respective weather condition, also for 120 minutes each. In total, this results in 1320 minutes with 11 different combinations. 
\subsection{Results}
\begin{figure}[h!]
	\centering
    % \captionsetup[subfigure]{labelformat=empty}
	\subfloat[lifespans]{\includegraphics[width=.46\linewidth]{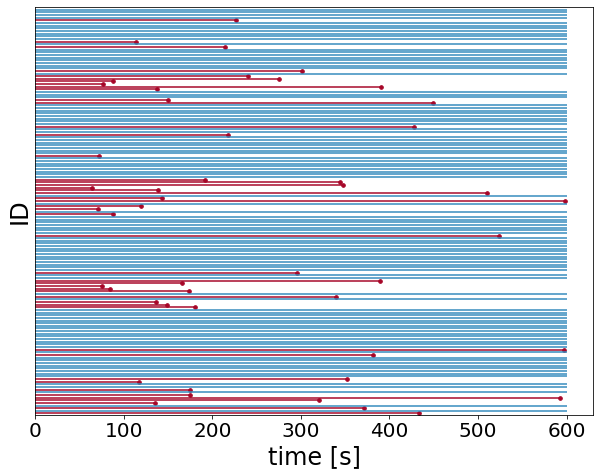}}~~
	\subfloat[Kaplan-Meier estimation]{\includegraphics[width=.49\linewidth]{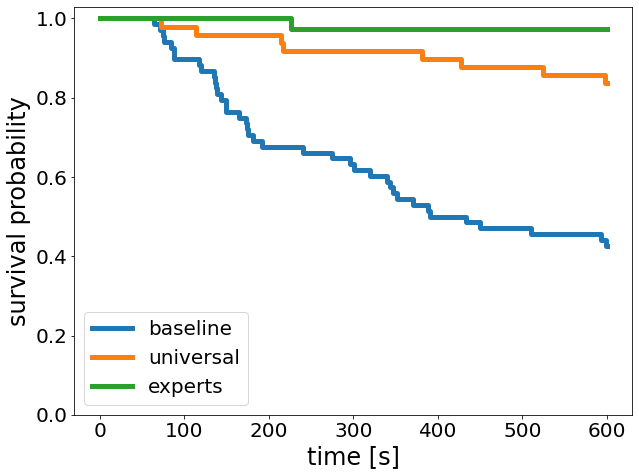}}~~
    \vspace{5pt}
    \caption[Lifespans and Kaplan-Meier estimation]{(a) Lifespans of all observations during the study. Red lines show the occurrence of a corner case, whereas blue lines are right-censored. The majority of the drives, approx. 70\%, did not lead to a corner case. (b) Kaplan-Meier estimation for all model types. The probability that no corner case occurs is highest in the expert models, followed by the universal model. The poor generalizability in bad weather provides that the survival probability in the base model decreases significantly over time.}
    \label{fig:lifelines_KM}	
\end{figure}

A total of $160$ drives with a maximum length of 600 seconds were performed. If no corner case occurs in this time, the drives are aborted so that we have a right-censored data point. Therefore, the number of rides per combination varies, as models in which a corner case appears more quickly can also be driven more frequently. The software used for survival analysis is lifelines~\cite{DavidsonPilon2019}. \Cref{tab:overview_cc} presents the total number of corner cases registered with respect to the trained model and the weather conditions driven. As we can see, there are barely corner cases in the expert models, which is why we group them together in their own model type, the experts type. All observations during the study are visualized in~\Cref{fig:lifelines_KM}(a). In total, we have 48 observations of corner cases that can be used for survival analysis. Furthermore, the two students drove $406.838$~km on the virtual streets of CARLA.

\begin{table}[h!]
	\begin{center}
     \caption[List of observed corner cases by model and tested weather condition]{List of all observed corner cases during weather campaign by model and tested weather condition.}
	\scalebox{.8}{

	\begin{tabular}{l|c|cccc}
		\hline
		\multirow{2}{*}{\textbf{model type}} & \multirow{2}{*}{\textbf{trained}} & \multicolumn{4}{c}{\textbf{tested}} \\ \cline{3-6} 
		 &  & \multicolumn{1}{c|}{clear} & \multicolumn{1}{c|}{rain} & \multicolumn{1}{c|}{fog} & night \\ \hline
		 \rowcolor[gray]{0.9}\textbf{baseline} & clear & \multicolumn{1}{c|}{4} & \multicolumn{1}{c|}{5} & \multicolumn{1}{c|}{13} & 17 \\ 
		\multirow{3}{*}{\textbf{experts}} & rain & \multicolumn{1}{c|}{-} & \multicolumn{1}{c|}{0} & \multicolumn{1}{c|}{-} & - \\ 
		 & fog & \multicolumn{1}{c|}{-} & \multicolumn{1}{c|}{-} & \multicolumn{1}{c|}{0} & - \\  
		 & night & \multicolumn{1}{c|}{-} & \multicolumn{1}{c|}{-} & \multicolumn{1}{c|}{-} & 1 \\ 
		 \rowcolor[gray]{0.9}\textbf{universal} & mix & \multicolumn{1}{c|}{1} & \multicolumn{1}{c|}{3} & \multicolumn{1}{c|}{1} & 3 \\ \hline
		\end{tabular}
}

\label{tab:overview_cc}
\end{center}
\end{table}

As a first step, we consider the plot for the Kaplan-Meier estimation in~\Cref{fig:lifelines_KM}(b) for the 3 model types \emph{baseline, universal} and \emph{experts}, which shows that the probability of a corner case occurring is lowest for the expert network, closely followed by the universal network. The baseline model seems to be very sensitive to different weather conditions, which is why there is only a survival probability of 63.24\% after 300 seconds and at the end of the observation period only 42.65\%.

\begin{figure}[h!]
	\centering
	\includegraphics[width=\linewidth]{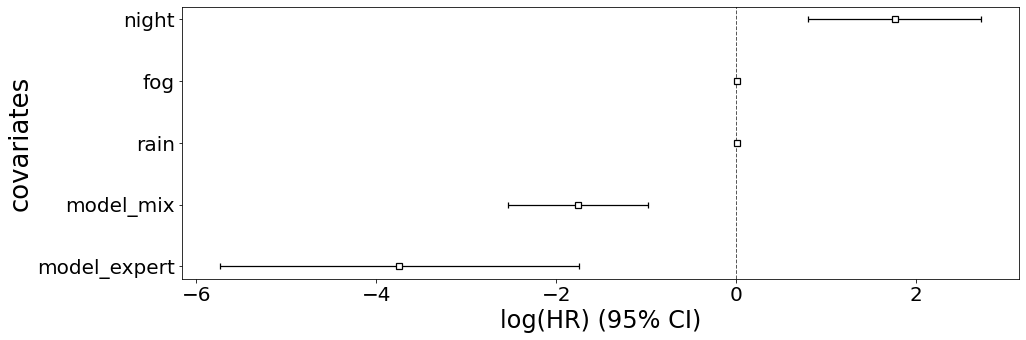}
	\caption[Hazard ratios]{The comparison of the hazard ratios shows that the night ensures that a corner case is more likely to occur. If an expert or universal model is used instead, a corner case occurs less frequently, which is also evident from the Kaplan-Meier estimate.}
	\label{fig:HR}
\end{figure}

We then use the Cox PH model to obtain the regression coefficients. For this, the input variables must first be preprocessed. For the weather parameter \emph{rain} the values can range from $70$ to $100$ and for \emph{fog} from $50$ to $100$. The parameter \emph{night} is assigned to a Boolean variable and the value 1 is set as soon as the sun position parameter ($\in [-90,90]$) is $<0$. Additionally, we distinguish on which model we are driving, for this we use also a Boolean variable and set a 1 for either the expert model or the universal model. 

\Cref{tab:coxPH} shows the evaluations of the Cox-PH. The analysis demonstrates that 3 covariates can be classified as significant, as their confidence interval is below 0.05. Fog is significant with 92\% and rain even only with 58\%. 

The hazard rate is calculated using the expert model as an example. Since this value is a boolean variable, it can be calculated as follows:
\begin{align}
	HR_{expert} = \frac{h_{expert=1}(t)}{h_{expert=0}(t)} =  0.02
\end{align}
Driving with an expert model reduces the hazard rate by 98\% with a low ranging confidence interval.

\begin{table}[h!]
	\begin{center}
    \caption[Cox PH model]{Cox PH model}
	\scalebox{.8}{
		\begin{tabular}{l|c|c|c}
			\hline
			\textbf{covariate} & \textbf{\begin{tabular}[c]{@{}c@{}}Hazard Ratio \\ HR\end{tabular}} & \textbf{\begin{tabular}[c]{@{}c@{}}95\% confidence interval \\ for the hazard ratio\end{tabular}} & \textbf{\begin{tabular}[c]{@{}c@{}}confidence level\\ p\end{tabular}} \\ \hline
			\rowcolor[gray]{0.9} rain & 1.01 & 0.99 - 1.02 & 0.42 \\ 
			fog & 1.01 & 1.00 - 1.02 & 0.08 \\ 
			\rowcolor[gray]{0.9} night & 5.83 & 2.23 - 15.22 & $<0.005$ \\ 
			experts & 0.02 & 0.00 - 0.17 & $<0.005$ \\ 
			\rowcolor[gray]{0.9} universal & 0.17 & 0.08 - 0.38 & $<0.005$ \\ \hline
			\end{tabular}
}

\label{tab:coxPH}
\end{center}
\end{table}

\begin{figure}[h!]
    \centering
	\subfloat[baseline]{\includegraphics[width=0.32\textwidth]  {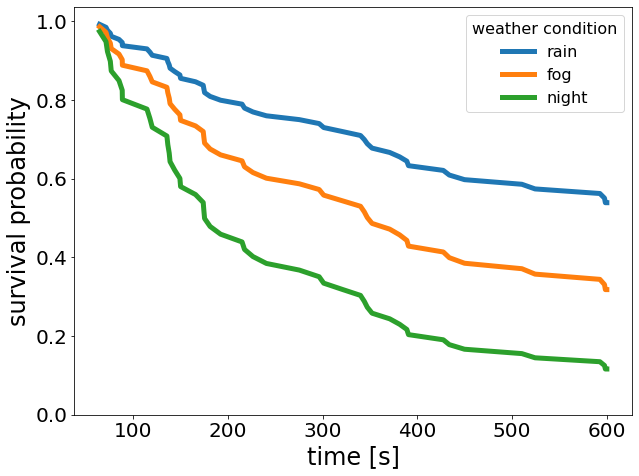}}~~
    \subfloat[universal]{\includegraphics[width=0.32\textwidth]	{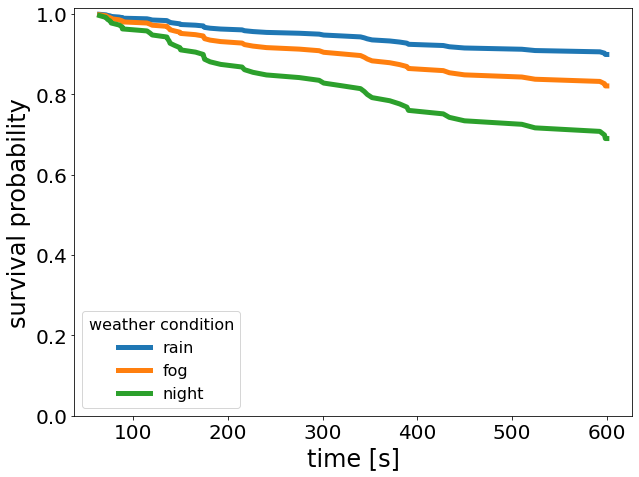}}~~
    \subfloat[experts]{\includegraphics[width=0.32\textwidth] {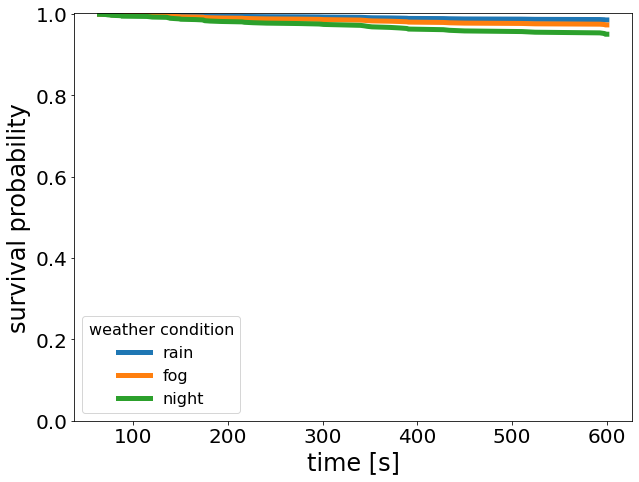}}~~
    \vspace{5pt}
    \caption[Survival probabilities for all models in different weather conditions]{The survival probabilities for rain, fog and night clearly show that the baseline model struggles with all bad weather settings. The universal model seems to be more robust, but the probability of survival at night also drops to 69\% at the end of the study, whereas the expert model assures a survival of 95\%.}
    \label{fig:experiments_cox}
\end{figure}

Next we have a closer look to the probabilities for all models in different weather conditions. \Cref{fig:experiments_cox} shows the performance for all models over time and for the weather conditions \emph{rain, fog, night}. The baseline model has the biggest problems when driving on unseen weather conditions, with the highest probability of a corner case occurring at \emph{night}. It also appears to be the most problematic for the universal and expert models, with significantly higher survival probabilities. The comparison between the universal and the expert models indicates that the latter perform noticeably better on their trained domains than the universal models. 

\section{\uppercase{Conclusion}}\label{sec:conclusion}
Due to the lack of explanation and transparency in the decision-making of today's AI algorithms, we developed an experimental setup that allows visualizing these decisions and thus allows a human driver to evaluate the driving situations while driving with the eyes of AI, and from this to extract data that includes safety-critical driving situations. Our self-developed test rig provides two human drivers controlling the ego vehicle in the virtual world of CARLA. The semantic driver receives the output of a semantic segmentation network in real-time, based on which she or he is supposed to navigate in the virtual world. The second driver takes the role of the driving instructor and intervenes in dangerous driving situations caused by misjudgments of the AI. We consider driver interventions by the safety driver as safety-critical corner cases which subsequently replaced part of the initial training data. We were able to show that targeted data enrichment with corner cases created with limited perception leads to improved pedestrian detection in critical situations. In addition, we continue the further development of AI by means of human risk perception to identify situations that are particularly important to humans and thus train the AI precisely where it is particularly challenged by a human perspective. 

The experimental setup with its components, the software used and the inference sensor have already been described in detail in~\cite{kowol22simulator}, as well as the proof that corner cases occur less frequently when they are generated by a driving simulator with weak perception and then used for training. 
Based on this, survival analysis was used to investigate whether universal models could be replaced by expert models trained for specific domains only, in order to save development time for the application. Although the validity of such a few data points must be treated with caution, a trend does seem to emerge, namely that the use of expert models indeed seems to be more appropriate, as an omniscient model has to find a balance to perform well in each domain. Therefore, it may be useful to focus on some basic data and add other models for special cases that are temporarily responsible for prediction. Examples of use would be driving in left-hand traffic or in snowy winter regions, so that an appropriately trained model could be used. It would also be conceivable to have a separate trained model for each country that may be used when crossing borders. This solution might be based on the vehicle's GPS coordinates and would not require an additional upstream classification network.

\section*{\uppercase{Acknowledgements}}

\noindent The research leading to these results is funded by the German Federal Ministry for Economic Affairs and Climate Action within the project KI Data Tooling under the grant number 19A20001E. We thank Matthias Rottmann for his productive support, Natalie Grabowsky and Ben Hamscher for driving the streets of CARLA.
\bibliographystyle{splncs04}
\bibliography{bibliography}

\end{document}